\documentclass[journal]{IEEEtran}

\usepackage{times}
\usepackage{amsmath,amssymb,amsfonts}
\usepackage{epsfig}
\usepackage{epstopdf}
\usepackage{graphicx}
\usepackage{subfigure}
\usepackage{color}
\usepackage{multirow,hhline}
\usepackage{url}
\usepackage{balance}
\usepackage{algorithm, algpseudocode, setspace}

\definecolor{darkgreen}{RGB}{0,160,0}

\newcommand{\ru}    {\rule{0mm}{4mm}}
\newcommand{\red}   {} 

\newcommand{\hd}    {\widehat{\delta}}
\newcommand{\td}    {\widetilde{\delta}}

\newcommand{{\hx}}  {\widehat{x}}

\newcommand{\minus} {\! - \!}
\newcommand{\plus}  {\! + \!}

\newcommand{\be}    {\begin{equation}}
\newcommand{\ee}    {\end{equation}}
\newcommand{\ag}[1] {{$\theta$ = {#1}$^o$}}

\newcommand{\chk}{\checkmark}

\newcommand{\NN}   {N\hspace{-0.6mm}N}

\begin{document}

\title{A PatchMatch-based Dense-field Algorithm \\ for Video Copy-Move Detection and Localization}
\author{Luca D'Amiano, Davide Cozzolino, Giovanni Poggi, and Luisa Verdoliva
\thanks{The authors are with the Dipartimento di Ingegneria Elettrica e delle Tecnologie dell'Informazione -- Universit{\`a} Federico II di Napoli -- Naples, ITALY,
e-mail: \{luca.damiano, davide.cozzolino, poggi, verdoliv\}@unina.it.}
}
\maketitle

\markboth{D'Amiano et al., A PatchMatch-based Dense-field Algorithm for Video Copy-Move Detection and Localization}
{D'Amiano \MakeLowercase{\textit{et al.}}: A PatchMatch-based Dense-field Algorithm \\ for Video Copy-Move Detection and Localization}

\begin{abstract}
We propose a new algorithm for the reliable detection and localization of video copy-move forgeries.
Discovering well crafted video copy-moves may be very difficult, especially when some uniform background is copied to occlude foreground objects.
To reliably detect both additive and occlusive copy-moves we use a dense-field approach, with invariant features that guarantee robustness to several post-processing operations.
To limit complexity, a suitable video-oriented version of PatchMatch is used, with a multiresolution search strategy, and a focus on volumes of interest.
Performance assessment relies on a new dataset, designed ad hoc, with realistic copy-moves and a wide variety of challenging situations.
Experimental results show the proposed method to detect and localize video copy-moves with good accuracy even in adverse conditions.
\end{abstract}


\section{Introduction}
\label{sec:intro}
Nowadays,
anyone can easily modify the appearance and content of digital images by means of powerful and easy-to-use editing tools such as Adobe Photoshop, Paintshop Pro or GIMP.
This is becoming increasingly true also for digital videos.
Powerful and widespread tools exist for video editing, like Adobe After Effects and Premiere Pro,
which allow users to perform a number of video manipulations.
Most of the times, these have the only purpose of improving the quality of videos or their appeal.
Sometimes, however, they are not so innocent,
aiming at falsifying evidence in court, perpetrating frauds or discrediting people.
Therefore, as happened in the last few years for still images,
there is an increasing interest in the scientific community towards the detection and localization of video forgeries \cite{Milani2012}.

These can be divided in whole-frame forgeries and object forgeries.
The first type of attack consists in deleting, inserting or replicating entire groups of frames.
Clearly, this action is quite simple to perform,
but not very flexible, and allows only for a limited set of manipulations.
Methods aimed at detecting such attacks
try to discover anomalies induced in the temporal structure of the encoded stream \cite{Stamm2012},
or other types of inconsistency,
like artifacts due to double encoding \cite{Wang2006, Gironi2014},
and irregularities in motion-compensated edges \cite{Su2009} or in the velocity field \cite{Wu2014}.
To detect whether a group of frames has been deleted
the use of {\it ad hoc} statistical features extracted from the motion residual has also been proposed \cite{Feng2014, Ravi2014}.

\begin{figure}[t!]
	\begin{tabular}{cc}
		\includegraphics[scale=0.154]{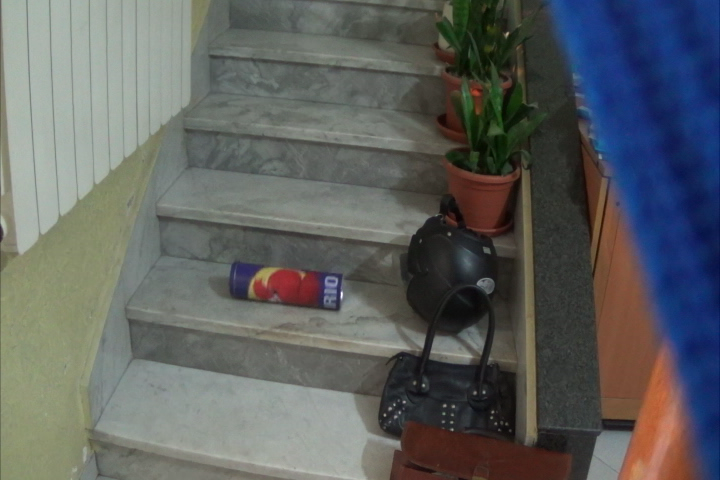} &
		\includegraphics[scale=0.154]{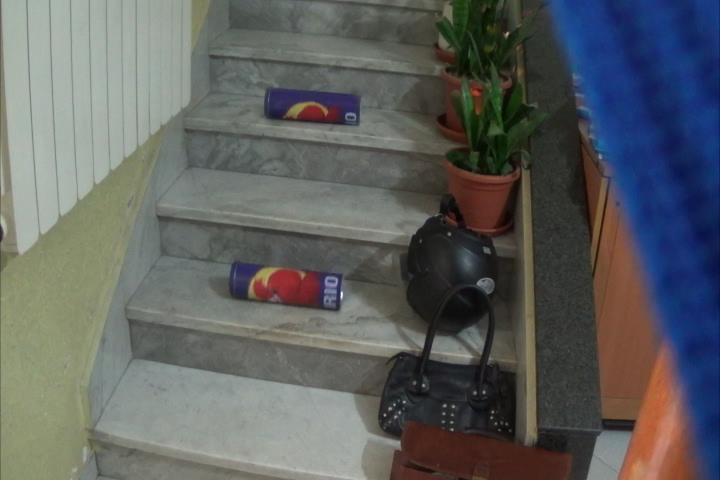} \\
		{\footnotesize (a) Original frame} &
		{\footnotesize (b) Additive forgery} \vspace{2mm}
        \\
		\includegraphics[scale=0.216]{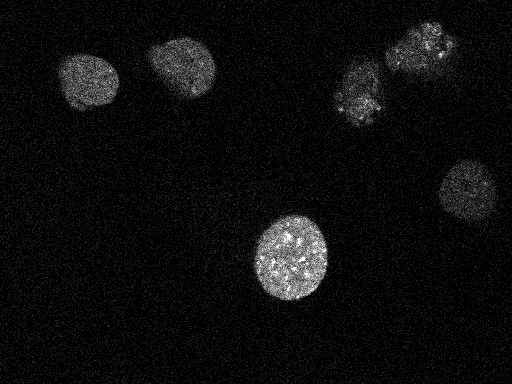} &
		\includegraphics[scale=0.216]{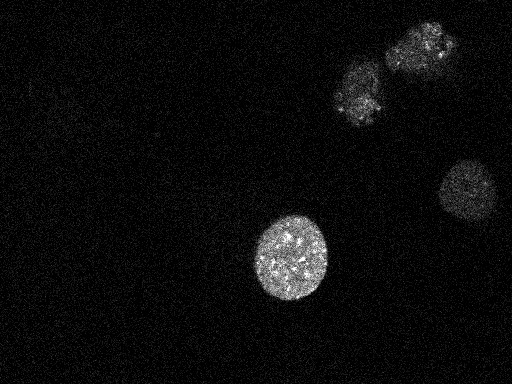} \\
		{\footnotesize (a) Original frame} &
		{\footnotesize (b) Occlusive forgery}
	\end{tabular}
	\caption{Additive (top) and occlusive (bottom) object copy-moves.
            Cell counting images, like those shown in the bottom, can be easily manipulated (https://ori.hhs.gov/) to commit scientific frauds.
            Detection may be quite difficult, especially in the occlusive case, due to the lack of salient keypoints.}
\label{fig:Object_Copymoves}
\end{figure}

Object forgeries, instead, concern the insertion or deletion of compact video objects.
For example, one might remove a person from a surveillance video by replacing it with suitable material taken from the same or other videos \cite{Wickramasuriya2005}.
Object forgeries are in general more sophisticated than whole-frame forgeries.
They are more difficult to perform, but allow for more flexible and subtle content modifications.
Moreover, if properly carried out \cite{Granados2012a,Granados2012b},
they can be quite challenging to detect, as they leave no obvious traces in the video temporal structure.


{\red
These attacks can be additive, when a video object of interest is inserted anew in the target video,
or occlusive, when an object is deleted from the video, typically through inpainting or by copying background over it.
Fig.\ref{fig:Object_Copymoves} shows examples of both situations.
Attacks can be also classified based on the object source, which may be the same video (copy-move), another video (splicing), or computer models (synthesis).
}

In recent years, only a few pioneering papers have addressed the detection of video object forgeries.
Coding-based methods have been proposed in \cite{Wang2009, Sun2012, Labartino2013, He2016}
where artifacts introduced by doubly-compressed MPEG videos are used as evidence of tampering.
An alternative approach relies on
detecting the camera ``fingerprint'' (camera PRNU pattern) as already done for images \cite{Chen2008, Chierchia2014}.
In \cite{Mondaini2007} the camcorder fingerprint is estimated on the first frames of the video and used to detect various types of attacks.
A similar idea is followed in \cite{Hsu2008, Chen2015, Davino2017} where
manipulations are discovered by extracting and analyzing some suitable features from the noise residues of consecutive frames.
In \cite{Kobayashi2010}, instead, the camera-dependent photon shot noise is used as an alternative to the camera fingerprint for static scenes.

Several papers have addressed video copy-move detection and localization.
In \cite{Wang2007} the correlation coefficient is used as a measure of similarity for detecting large copy-moved blocks,
while \cite{Bestagini2013} extends the same method to use spatio-temporal blocks.
However, this approach works well only if the cloned area is relatively large and has not been subject to subsequent post-processing.
Performance drops in the presence of compression, blurring, geometric transformations and change of intensity.
This is not immaterial,
as these operations are often needed to make the forgery more realistic,and can be even enacted on purpose by a skilled forger to fool forensic tools.
For what concern localization, instead,
\cite{Liao2013} addresses the case of whole frames inserted in the video,
while \cite{Subramanyam2012} requires very strong hypotheses,
like considering only rigid copy-moves or supposing that the copy is performed between consecutive frames.

{\red
It should be realized that a carefully crafted copy-move may be very hard to discover by means of statistical approaches,
because the copied object has the same statistics as the background, unless it is rotated or resized.
In addition, occlusive copy-moves, based on the copy of background areas, do not offer visual clues or salient keypoints (see again Fig.\ref{fig:Object_Copymoves})
which enable their discovery.
Finally, a clever attacker may enact further expedients to confuse matching-based methods, like playing the video backwards.

In this paper we propose a new technique for the detection and localization of copy-move video forgeries.
First, suitable features are computed, invariant to various spatial, temporal, and intensity transformations.
The features are computed {\em densely} on a spatio-temporal grid, rather than at salient keypoints, which allows us to detect not only additive but also occlusive forgeries.
Afterwards, a nearest-neighbor field (NNF) is built, connecting each feature with its best-matching.
To this end, we use an {\it ad hoc} video-oriented version \cite{Cozzolino2015,Damiano2015} of PatchMatch \cite{Barnes2009,Barnes2010}, exploiting the inherent coherency of the NNF to reduce search complexity.
Finally, the NNF is post-processed to single out areas with coherent spatio-temporal displacement as candidate copy-moves.

This paper extends our conference paper \cite{Damiano2015}, based in turn on previous work on image copy-move detection \cite{Cozzolino2014,Cozzolino2015}.
However, with respect to \cite{Damiano2015} we
\begin{enumerate}
\item   define a new flip-invariant version of our features;
\item   introduce of new criterion in the post-processing to tell apart copy-moves from false matches;
\item   design a fast version of the detector, based on multi-scale processing and parallel implementation;
\item   contribute a new specific dataset with realistic copy-moves, both additive and occlusive;
\item   carry out a thorough performance analysis taking into account the most challenging situations of interest.
\end{enumerate}
}

The rest of the paper is organized as follows.
Section 2 provides the necessary background information, reviewing ideas and tools for still-image copy-move detection, with special focus on our previous work \cite{Cozzolino2015} and on the PatchMatch algorithm.
Section 3 describes in detail the proposed technique.
Then, in Section 4 we discuss the results of a number of experiments carried out to test the proposed algorithm in various operative conditions.
Finally, Section 5 draws conclusions.

\section{Background}

In the last few years,
a large number of techniques have been proposed for the detection and localization of copy-move forgeries in digital images \cite{Christlein2012}.
Virtually all such techniques comprise three major steps:
{\em   i)} feature extraction,
{\em  ii)} matching, and
{\em iii)} post-processing.
In the first step a suitable feature is associated with each pixel of interest.
Based on such features, each pixel is then linked with its best match over the image, generating a field of offsets.
Finally, this field is processed to single out regularities which point at possible copy-moves.

Some techniques, e.g., \cite{Pan2010, Amerini2011, Zhao2013},
operate only on a small set of salient keypoints,
characterized through well-known local descriptors, such as SIFT or LBP.
This approach is computationally efficient,
but fails completely if no keypoint is associated with the forgery, as in the common case of occlusive copy-moves over a smooth background \cite{Christlein2012, Cozzolino2015}.

Techniques based on dense sampling are much more reliable.
Their main issue is complexity, since all pixels are involved in the three phases of feature extraction, matching, and post-processing.
To reduce computation, compact features are extracted, typically through some transforms,
like DCT \cite{Fridrich2003}, wavelet \cite{Muhammada2012}, PCA \cite{Mahdian2007} or SVD \cite{Zhao2013a}.
By so doing, a good robustness is also obtained with respect to intensity distortions, originated for example by JPEG compression or blurring.
Instead, to deal with geometric distortions due to rotated or rescaled copy-moves, specific invariant features are needed.
The Zernike moments and the polar sine and cosine transforms have been used \cite{Ryu2013, Li2013, Li2014} to obtain rotation invariance,
while for scale-invariance the Fourier-Mellin Transform with log-polar sampling has been considered \cite{Bayram2009, Wu2011}.

Featuring, however, is only part of the problem.
The bulk of complexity for dense-field techniques resides in the matching phase.
Barring the trivial case of identical copy-moves, where simple lexicographic sorting can be applied,
exhaustive search of the best matching (nearest neighbor) feature is prohibitively complex,
and faster techniques must be devised to produce the offset field in a reasonable time.
To this end, approximate search strategies have been used,
such as kd-tree search, in \cite{Langille2006, Christlein2012},
or locality sensitive hashing, in \cite{Ryu2013, Li2013}.
Nonetheless, computing the nearest-neighbor field keeps being too slow for the large images generated by today's cameras.
A much better result can be obtained, however, by exploiting the strong regularity exhibited by the NNFs of natural images,
where similar offsets are often associated with neighboring pixels.
This is done in \cite{Cozzolino2014} and \cite{Cozzolino2015},
where the offset field is computed by means of a suitably modified version of PatchMatch \cite{Barnes2009,Barnes2010},
a fast randomized search technique specifically tailored to the properties of images.
In the following, we provide first a brief description of PatchMatch
and then of the technique proposed in \cite{Cozzolino2015} which is the starting point for the current video-oriented proposal.

\subsection{PatchMatch}

Let
$
    I = \{I(s) \in R^K, s \in \Omega\}
$
be an image defined over a regular rectangular grid $\Omega$.
With each pixel $s$ we associate a feature vector, $f(s)$, which describes the image patch centered on $s$.
Given a suitable measure of distance between features, $D(f',f'')$,
we define the nearest neighbor of $s$ as the pixel, $s' \in \Omega$, which minimizes the feature distance w.r.t. $s$ over the whole image
\begin{equation}
    {\rm NN}(s) = \arg\min_{s' \in \Omega} D(f(s),f(s'))
    \label{eq:nearest_neighbor}
\end{equation}
Rather than the nearest-neighbor field (NNF) itself,
in the following we will consider the equivalent offset field, with the offset defined as $\delta(s)={\rm NN}(s)-s$.

PatchMatch is a randomized iterative algorithm for NNF computation.
As all iterative algorithms, convergence to the desired solution is much faster in the presence of a good initial guess.
With images, however, such a good guess is easily obtained,
because their NNFs are typically constant or linearly varying over large areas, as a consequence of image smoothness, and hence highly predictable.
Given this core idea, PatchMatch is easily understood.
Following a random initialization, the two phases of offset prediction and random search alternate until convergence.

\vspace{2mm} \noindent {\em Initialization.}
The offset field is initialized at random, as
$\delta(s) = U(s)-s$.
where $U(s)$ is a bi-dimensional random variable, uniform over the image support $\Omega$.
In copy-move search, we set the additional constraint that matches should be reasonably far from the target,
excluding offsets smaller than a given threshold.
Most of the initial offsets are useless, but a certain number will be optimal or near-optimal.
These are quickly diffused to the rest of the image in the propagation phase.

\vspace{2mm} \noindent {\em Propagation.}
In this step, the image is raster scanned top-down and left-to-right (with scanning order reversed at every other iteration),
and for each pixel $s$ the current offset is updated as
\begin{equation}
    \delta(s) = \arg\min_{\phi \in \Delta^P(s)} D(f(s),f(s+\phi))
    \label{eq:propagation}
\end{equation}
where $\Delta^P(s)=\{\delta(s),\delta(s^r),\delta(s^c)\}$,
and $s^r$ and $s^c$ are the pixels preceding $s$, in the scanning order, along rows and columns, respectively.
Therefore, the algorithm uses the offset of nearby pixels as alternative estimates of the current offset, and selects the best one.
If a good offset is available for a given pixel of a region with constant offset,
this will very quickly propagate to the whole region.

\vspace{2mm} \noindent {\em Random search.}
To avoid getting trapped in bad local minima,
after each propagation step a random search step follows, based on a random sampling of the current offset field.
The candidate offsets ${\delta_i(s), i=1,\ldots,L}$ are chosen as
$\delta_i(s) = \delta(s) + R_i $
where $R_i$ is a bi-dimensional random variable, uniform over a square grid of radius $2^{i-1}$, excluding the origin.
In practice, most of these new candidates are pretty close to $\delta(s)$, but large differences are also allowed, with small probability.
Given the rare sampling, only a few new candidates are eventually selected.
The random-search updating reads therefore as
\begin{equation}
    \delta(s) = \arg\min_{\phi \in \Delta^R(s)} D(f(s),f(s+\phi))
    \label{eq:random_search}
\end{equation}
where $\Delta^R(s)=\{\delta(s), \delta_1(s), \ldots, \delta_L(s)\}$.

Experiments \cite{Barnes2009} show that typically PatchMatch converges to a near-optimal NNF in less than 10 iteration.

\subsection{A PatchMatch-based technique for still-image copy-move detection}

In \cite{Cozzolino2015} we proposed a new technique for copy-move detection and localization in still images.
Thanks to the use of rotation-invariant and robust features, copy-moves are reliably detected even in the presence of various forms of intensity and geometric distortion.
Efficiency is ensured by using a suitably modified version of PatchMatch for the offset field computation
and a fast {\it ad hoc} post-processing to remove false matches.

Let $I(\rho, \theta)$ be the input image in polar coordinates, with $\rho \in [0,\infty]$ and $\theta \in [0,2\pi]$, and let
\begin{equation}
    K_{n,m}(\rho,\theta) =  R_{n,m}(\rho) \frac{1}{\sqrt{2\pi}}e^{jm\theta}
\end{equation}
be a kernel function obtained as the product of a radial profile $R_{n,m}(\rho)$ and a circular harmonic.
By projecting the image over the kernel we obtain the feature
\begin{equation}
    f(n,m) = \int_0^\infty \! \rho R_{n,m}^*(\rho) \times \left[ \frac{1}{\sqrt{2\pi}}\int_0^{2\pi} I(\rho,\theta) e^{-jm\theta}d\theta \right] \, d\rho\
\end{equation}
By choosing the Zernike orthonormal radial functions \cite{Teague1980}
$f(n,m)$ turns out to be the Zernike moment of order $(n,m)$ of the image.
Note that the integral in square brackets is the Fourier series of $I(\rho,\theta)$ along the angle coordinate, and its magnitude is invariant to rotations of the image $I$.
Therefore, by selecting as features the magnitude of Zernike moments we guarantee rotation invariance.
In addition, if only a few low-order moments are used, a compact feature vector is obtained, robust to intensity distortions, which are mostly of high-pass nature.

{\red

\begin{figure}[t]
\centering
\begin{minipage}[c]{.32\linewidth} \centerline{\epsfig{figure=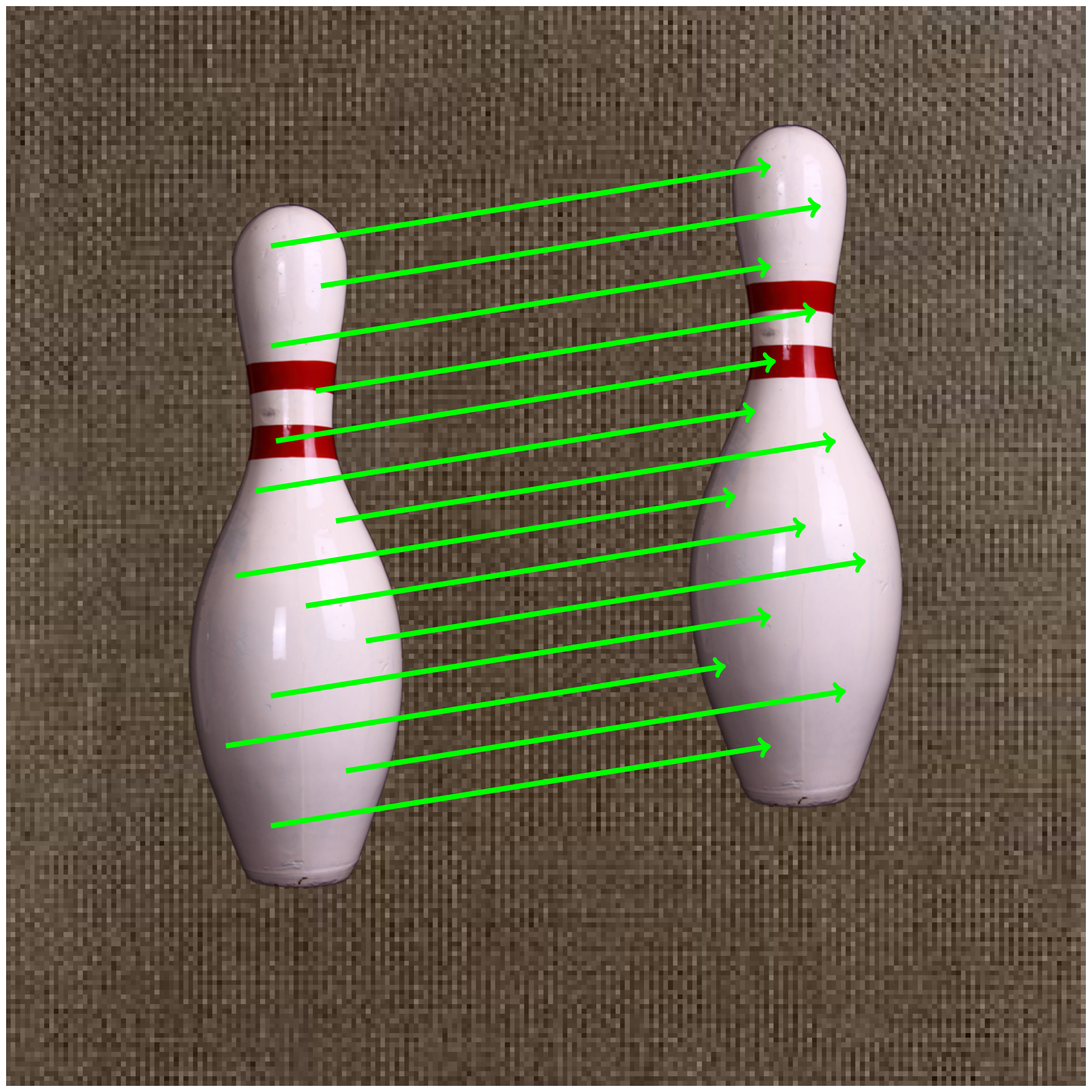, width=2.8cm}} \end{minipage} \hfill
\begin{minipage}[c]{.32\linewidth} \centerline{\epsfig{figure=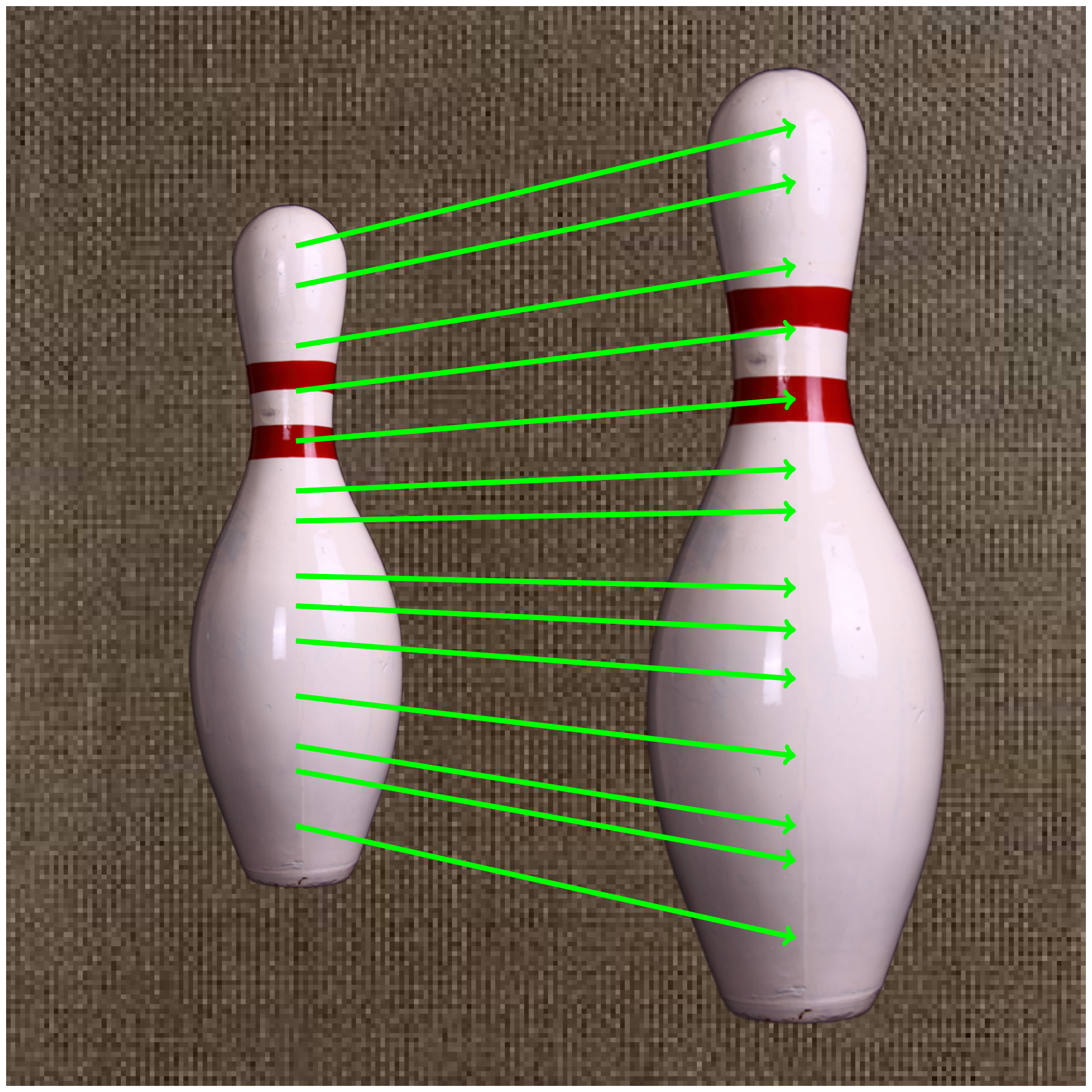, width=2.8cm}} \end{minipage} \hfill
\begin{minipage}[c]{.32\linewidth} \centerline{\epsfig{figure=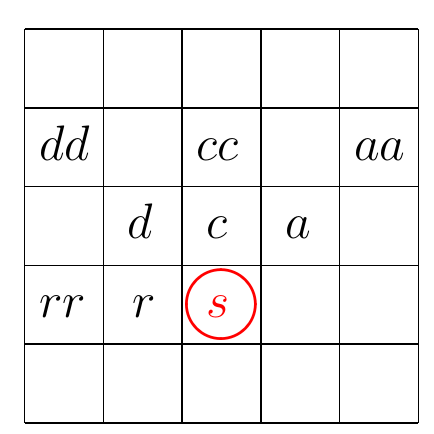, width=2.8cm}} \end{minipage} \hfill
\caption{Modified PatchMatch.
With rigid copy-moves (left) clones are connected by a constant offset field.
In the presence of rotation and/or resizing (center) clones are connected by a linearly varying offset field.
PatchMatch (right) uses zero-order predictors of the offset,
based on neighboring pixels $(r,c,d,a)$ on the same row, column, or diagonal as the target $(s)$.
The modified version uses also first-order predictors, involving neighbors farther apart $(rr,cc,dd,aa)$ so as to follow linear variations.}
\label{fig:Modified_PatchMatch}
\end{figure}

To compute the offset field efficiently we resort to PatchMatch.
However, the basic version of the algorithm is designed for patchwise constant offset fields, a model appropriate for rigid copy-moves, as in Fig.\ref{fig:Modified_PatchMatch}(a),
while rotated and resized copy-moves give rise to linearly varying offsets, as in Fig.\ref{fig:Modified_PatchMatch}(b).
A generalized version of PatchMatch was proposed in \cite{Barnes2010} to deal with this problem.
Unfortunately,
it works only on image patches (not compact features) and is significantly more complex than the basic version.
A much simpler modification was proposed in \cite{Cozzolino2014},
adding first-order predictors to the zero-order predictors used in PatchMatch, so as to deal effectively also with linear offset fields.
With reference to Fig.\ref{fig:Modified_PatchMatch}(c),
a zero order prediction of the offset $\delta(s)$ at site $s$ is given by
\begin{equation}
    \td^{0x}(s) =  \delta(x), \hspace{6mm} x \in \{r,d,c,a\}
\end{equation}
that is, the offset is predicted as being equal to the offset of the neighbor on the same row, column, diagonal or antidiagonal.
Adding first-order predictors
\begin{equation}
    \td^{1x}(s) = 2\delta(x)-\delta(xx)
\end{equation}
we take into account linear variations of the offset along the same four directions.
Eventually, we obtain the enlarged set of predicted offsets
\begin{eqnarray}
    \Delta^P(s) & = & \{\delta(s), \td^{0r}(s), \td^{0d}(s), \td^{0c}(s), \td^{0a}(s), \nonumber \\
                &   &              \td^{1r}(s), \td^{1d}(s), \td^{1c}(s), \td^{1a}(s)\}
\end{eqnarray}
which are used in the propagation phase to perform the search of equation (2).
}

Finally, to take full advantage of PatchMatch's efficiency, the post-processing phase must equally fast.
With this aim, an {\it ad hoc} post-processing was implemented, called dense linear fitting (DLF).
An affine model is fit locally to each point of the offset field, with parameters estimated from the data themselves.
The fitting is typically good in correspondence of a copy-moved regions,
where the offset field is either constant (for plain copy-moves) or linearly varying (in the presence of rotations or resizing).
On the contrary, in pristine areas of the image, with a more chaotic field, a worse fit is typically observed.
Therefore, by looking for large areas with low fitting error, copy-moves can be reliably detected.
The fitting procedure is very fast, as it only requires a few linear filtering and products per pixel.

\section{Proposed technique}

To address the detection and localization of video copy-moves we extend the technique proposed in \cite{Cozzolino2015} for still images which,
thanks to its dense-field approach, is effective with both additive and occlusive copy-moves.
Therefore, our proposed technique comprises the three usual phases of
\begin{itemize}
\item   (dense) feature extraction;
\item   feature matching;
\item   post-processing of the nearest-neighbor field.
\end{itemize}
To take advantage of our past experience, we move from the same basic tools used in \cite{Cozzolino2015},
namely, Zernike moments, PatchMatch, and Dense Linear Fitting, respectively.
However, going from still images to videos, a number of new issues emerge, that must be addressed specifically.
Features must be adapted to ensure the necessary robustness to both temporal and spatial distortions;
PatchMatch itself must be adapted to deal efficiently with a video source;
the post-processing must be tailored to take care of the large number of false hits arising naturally in a highly redundant source.
Last, but not least, computational efficiency must be pursued.
At standard frame-rates, a minute of video corresponds to about 1500 frames
and complexity may soon become unmanageable even for short YouTube\textsuperscript{TM} videos.
All these issues are addressed in the following subsections.

\subsection{Features}

Building upon the still-image copy-move detector of \cite{Cozzolino2015}
we begin by associating with each pixel a feature vector composed by the Zernike moments computed on a polar grid centered on the target.
Dealing with a video source, however, we have the opportunity to extract features from 3D rather than 2D patches.
With this choice a more expressive feature is obtained, accounting also for informative temporal changes.
On the down side, 3D patches may be less effective with forgeries of very short duration or with fast moving objects,
because many 3D patches would include both pristine and copy-moved regions, making it difficult to find the correct match.
Moreover, they are more fragile with respect to temporal distortions, such as flipping or temporal down/up-sampling.
Since both solutions (3D-patch or 2D-patch based features) have pros and cons, we will test them both in the experiments.
For the second case, however, we define a flip-invariant feature so as to be robust to the simplest form of tampering in the temporal dimension.

{\red
Specifically,
let $f(s,t,n,m)$ be a generic feature associated with the $t$-th frame of the video, for spatial location $s$ and Zernike moment $(n,m)$.
Then, the 2D-patch feature vectors used in the algorithm, or 2D features for short, will be defined as
\begin{equation}
    {\bf f}^{\rm 2D}(s,t) = \{f(s,t,n,m), \; (n,m)\in {\cal F}^{\rm 2D} \}
\end{equation}
where ${\cal F}^{\rm 2D}$ identifies a subset of all Zernike moments.
This subset should be as small as possible to limit complexity and memory problems in the algorithm,
while including sufficient discriminative information on the patch.
Similarly,
we could define 3D-features as
\begin{equation}
    {\bf f}^{\rm 3D}(s,t) = \{f(s,t\plus\tau,n,m), \; |\tau| \leq T, (n,m)\in {\cal F}^{\rm 3D} \}
\end{equation}
where Zernike moments from 2$T$+1 consecutive frames are taken,
and ${\cal F}^{\rm 3D}$ identifies a new subset of Zernike moments, smaller than before to limit feature size.
This latter feature, however, is not flip invariant, and would not allow the detection of clones played backwards in time.
Therefore, to improve robustness to malicious attacks, we modify it as
\begin{equation}
    {\bf f}^{\rm 3D,FI}(s,t) = \{g(s,t\plus\tau,n,m), \; |\tau| \leq T, (n,m)\in {\cal F}^{\rm 3D} \}
\end{equation}
where, displaying only the dependence on the temporal index,
\begin{equation}
    g(t\plus\tau) = \left\{ \begin{array}{ll}
	       \frac{1}{\sqrt{2}} | f(t\plus\tau) + f(t\minus\tau)| & \tau > 0 \rule{0mm}{5mm} \\
                              | f(t) |                          & \tau = 0 \rule{0mm}{5mm} \\
	       \frac{1}{\sqrt{2}} | f(t\plus\tau) - f(t\minus\tau)| & \tau < 0 \rule{0mm}{5mm}
	\end{array} \right.
\label{eq:evenodd_transform}
\end{equation}
The even-odd transform of Eq.(\ref{eq:evenodd_transform}) guarantees the desired flip invariance.
Note that the transform is applied on the Zernike moments, not their absolute values.
By so doing, the ${\bf f}^{\rm 3D,FI}$ vectors are invariant to spatial rotations of the whole 3D patch.
Taking the absolute value of the moments before the even-odd transform would enforce invariance with respect to different rotations for each frame.
This is a useless property for our problem, since temporal and spatial manipulations do not interfere with one another.
We also note explicitly that the above formulas refer to a 3D patch with an odd number of frames, similar formulas apply for the even number case.
}

\begin{figure}[t]
\centering
\begin{minipage}[c]{.12\linewidth} {\color{white} .} \end{minipage} \hfill
\begin{minipage}[c]{.32\linewidth} \centerline{\epsfig{figure=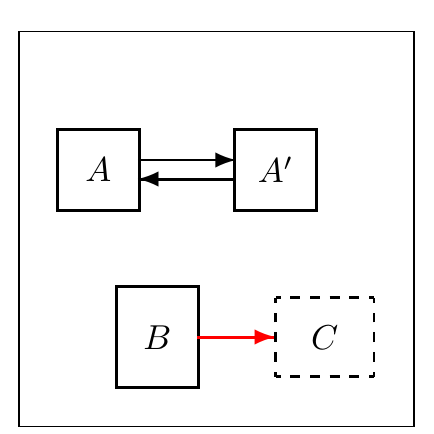, width=2.8cm}} \end{minipage} \hfill
\begin{minipage}[c]{.32\linewidth} \centerline{\epsfig{figure=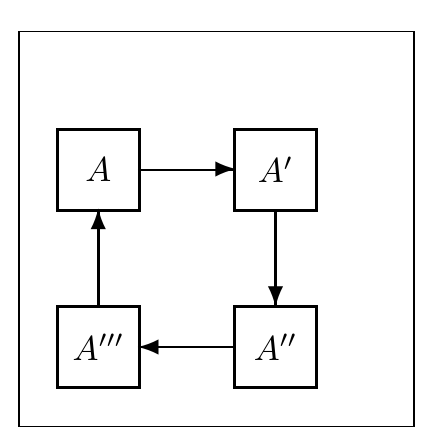, width=2.8cm}} \end{minipage} \hfill
\begin{minipage}[c]{.12\linewidth} {\color{white} .} \end{minipage} \hfill
\caption{Left:removing random false matches. The preliminary detection map $M^{\rm DLF}$ includes two clones, $A, A'$, and a false match $B$ pointing to region $C$ not in the map.
Since $B$ does not point to a region in the map it is eventually removed.
Right: multiple clones. The four clones, $A, A', A'', A'''$ all belong to the preliminary map.
Even though no two regions match one another, they all match regions in the map, so they are all kept.}
\label{fig:Post_Processing}
\end{figure}

\begin{figure*}[t]
	\centering
    \unitlength=1.3mm
	\begin{picture}(160,50)(-000,+002)
	
	\put(008,042){\vector(+1,+0){12}}
	\put(012,043){\small $V$}
	\put(020,039){\framebox(18,06){\small Featuring}}
	\put(038,042){\vector(+1,+0){12}}
	\put(042,043){\small $F^{0}$}
	\put(050,039){\framebox(06,06){$\downarrow$}}
	\put(056,042){\vector(+1,+0){42}}
	\put(074,043){\small $F^{1}$}
	\put(098,039){\framebox(06,06){$\downarrow$}}
	
	\put(104,042){\line(+1,+0){12}}
	\put(116,042){\vector(+0,-1){12}}
	\put(112,043){\small $F^{2}$}
	\put(118,039){\vector(+0,-1){09}}
	\put(119,034){\small $F^{0}$}
	\put(107,024){\framebox(18,06){\small PatchMatch}}
	\put(106,013){\small $\NN^{2}$}
	\put(116,024){\line(+0,-1){12}}
	\put(116,012){\vector(-1,+0){12}}
	\put(098,009){\framebox(06,06){$\uparrow$}}
	
	\put(098,012){\vector(-1,+0){12}}
	\put(089,013){\small $\NN^{1}_0$}
	\put(077,042){\vector(+0,-1){27}}
	\put(079,024){\vector(+0,-1){09}}
	\put(080,019){\small $F^{0}$}
	\put(068,009){\framebox(18,06){\small CMD}}
	\put(068,012){\vector(-1,+0){12}}
	\put(059,013){\small $\NN^{1}$}
	\put(074,015){\line(+0,+1){9}}
	\put(074,024){\vector(-1,+0){18}}
	\put(060,025){\small $M^{1}$}
	
	\put(050,021){\framebox(06,06){$\uparrow$}}
	\put(050,024){\line(-1,+0){18}}
	\put(032,024){\vector(+0,-1){09}}
	\put(040,025){\small VoI}
	
	\put(050,009){\framebox(06,06){$\uparrow$}}
	\put(050,012){\vector(-1,+0){12}}
	\put(040,013){\small $\NN^{0}_0$}
	\put(029,039){\vector(+0,-1){24}}
	\put(020,009){\framebox(18,06){\small CMD}}
	\put(020,012){\vector(-1,+0){12}}
	\put(012,013){\small $M$}
		
	\put(024,001){\small {\it full resolution}}
	\put(063,001){\small {\it $S  \!\times\!S  $ lower resolution}}
	\put(107,001){\small {\it $S^2\!\times\!S^2$ lower res.}}
	
	\multiput(053,000)(000,002){04}{\line(+0,+1){1}}
	\multiput(053,029)(000,002){04}{\line(+0,+1){1}}
	\multiput(053,048)(000,002){02}{\line(+0,+1){1}}
	
	\multiput(101,000)(000,002){04}{\line(+0,+1){1}}
	\multiput(101,017)(000,002){10}{\line(+0,+1){1}}
	\multiput(101,048)(000,002){02}{\line(+0,+1){1}}
	
	\end{picture}
\label{fig:Block_Diagram}
\caption{Block diagram of the proposed video copy-move detector with multi-resolution processing.
The high-resolution field of features $F^0$ is extracted from the original video, $V$.
This field is then downsampled twice to obtain fields  $F^1$ and $F^2$.
At level 2 (lowest resolution) PatchMatch works on $F^2$ and $F^0$ to provide the NN field $\NN^2$.
This is upsampled to become the initial NN field at level 1, $\NN^1_0$.
At level 1, the copy-move detector (CMD) works on $F^1$ and $F^0$ to refine $\NN^1_0$ to $\NN^1$,
and to extract the detection map $M^1$ by applying the post-processing.
Copy-moved objects are detected in this level, but their shape can be recovered more precisely at level 0.
So $M^1$ is upsampled to define the volume of interest (VoI) and $\NN^1$ is upsampled to become the initial NN field at level 0, $\NN^0_0$.
At level 0, the copy-move detector works on $F^0$, limited only to the VoI, to extract the final output, the detection map $M^0=M$.
}	
\end{figure*}
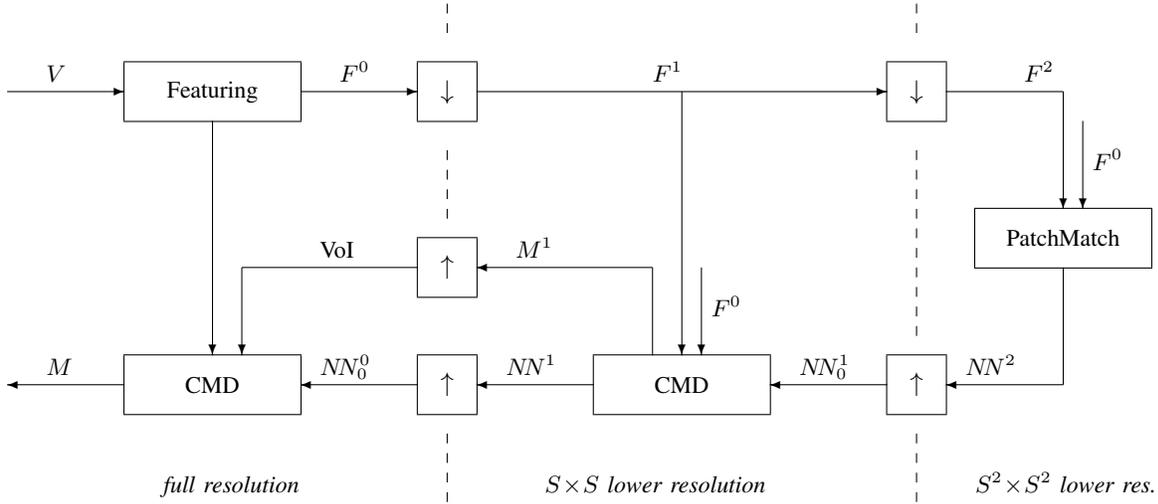


\subsection{Adapting PatchMatch to video}
To take advantage of the features' invariance to rotations, we use the modified version of PatchMatch proposed in \cite{Cozzolino2015}.
Then, to deal with a video source,
we adopt some further straightforward modifications, originally developed in \cite{Damiano2015}.
First of all,
while keeping the general structure of the algorithm, we include further predictors to take into account the temporal direction,
and in particular the zero-order and first-order predictors along frames $\hd^{0f}(s)$ and $\hd^{1f}(s)$,
defined like in equations (6) and (7).
Using first-order prediction along time allows one to deal also with subsampling [upsampling] in the temporal direction, corresponding to a change of speed in moving objects.
In addition, the random search step is also modified to sample the whole 3D space, testing offsets taken at random in the whole datacube.
Despite the increased size of the source, with a large number of frames, the same number of candidates is used as for still images.
Since this procedures is repeated for all pixels in the video, a large number of near-optimal offsets are sampled anyway, and then propagated to the whole video.

\subsection{post-processing}

The NNF produced by PatchMatch is subsequently processed by the Dense Linear Fitting algorithm
which singles out regions with coherent offset field and associates a binary detection map $M^{\rm DLF}$ with the video.
However, since similar pristine regions abound in images, some false alarms may also arise at this point.
In \cite{Cozzolino2015} these were largely eliminated through morphological filtering.
When dealing with videos, however, this problem becomes much more relevant because subjects, and especially background areas,
appear almost identical in many subsequent frames, giving rise to a large number of false alarms.
Standard morphological filtering cannot solve the problem anymore.
We therefore add a further control, working always on the NNF to keep high efficiency.

The inspiring principle is that true clones should match both ways, that is
\begin{equation}
    (s,t)+\delta(s,t) = (s',t') \;\Leftrightarrow\; (s',t')+\delta(s',t') = (s,t)
\end{equation}
Points for which this condition does not hold may be random matches rather than corresponding points of a copy-move.
Actually, this is too strict a condition to enforce, since small deviations from this rule apply also to actual copy-moves.
In addition, with multiple clones, the very same principle weakens, as points may exhibit a circular matching.
Therefore we use a weaker but still effective constraint,
requiring simply that regions matching through the NNF all belong to preliminary detection map $M^{\rm DLF}$.
This is rarely the case for false matches, while it happens with near certainty for actual copy-moves.
{\red
Pictorial examples of the application of these rules are shown in Fig.\ref{fig:Post_Processing} with reference to a 2D geometry.
}

\subsection{Managing complexity}

The overall complexity of the basic algorithm of \cite{Cozzolino2015} is clearly dominated by the matching phase,
accounting for about 75\% of the total CPU-time,
with the computation of features taking another 15\%, and the post-processing phase the remaining 10\%.
In the case of video, these proportions are not likely to change much.
Therefore all efforts should be devoted to further reduce the cost of computing the NN field.

Using PatchMatch, the complexity of computing the NN field, measured in number of multiplications, can be approximated as
\begin{equation}
    C = N_{it}N(C_P+C_R)F
\end{equation}
where
$N_{it}$ is the number of iterations of PatchMatch,
$N$ the number of pixels in the video,
$C_P$ and $C_R$ are the number of candidate offset tested in the propagation and random search phases, respectively,
and $F$ is the feature length.
$N$ in its turn is the product of frame size and number of frames.
Using typical values,
that is, frames of 0.5 Mpixels, 8 iterations of PatchMatch, with 10 candidates tested in each phase, and features of length 10,
the overall complexity is $8\times 10^8$ multiplications per frame.
At 25 frames/second, this represents a huge computational burden, even for short videos.

The only effective way to reduce significantly this burden is through subsampling.
Indeed, keypoint-based methods use exactly this strategy, computing matches only for some sparse keypoints.
As we follow a dense-field approach, we perform instead a regular $S \times S$ subsampling (that is, we take one every $S$-th pixel along rows and columns).
Note that features are always computed on the original frames {\em before} subsampling,
therefore they represent patches observed at the native resolution, say level-0, with no loss of information.
Also, no subsampling is performed along the temporal direction.

By working at level-1 resolution (that is, after subsampling) PatchMatch complexity reduces by a factor $S^2$, approximately, if all other parameters are kept fixed.
Moreover, with a moderate subsampling step, we expect to keep detecting at level-1 most, if not all, the copy-moves detectable at level-0.
Of course, there is a loss in spatial accuracy.
However, this can be largely recovered by upsampling the NN field back at level-0,
and running a few iterations of PatchMatch to propagate the correct offsets.
Since detection has been already performed at level 1,
at level 0 PatchMatch is applied only to the volumes of interest (VoI), namely the frames where copy-moves have been detected,
while the random search phase is skipped altogether.
With these simple modifications the processing cost at level 0 becomes fully manageable.
In particular, since it is quite unlikely that a copy-move lasts for more than a few seconds, the use of VoI alone is already very effective.

\begin{algorithm}
	\footnotesize
	\begin{spacing}{1.25}
		\begin{algorithmic}[1]
			\Require $V$ \ru                                        \Comment input video
			\Ensure  $M$                                            \Comment output detection map
			\State $F^{0}$ = FeatureExtract($V$)                    \Comment will work on features from now on
			\State $F^{1}$ = $F^{0} \downarrow S$                   \Comment $S\times S$ downsampling
			\State $F^{2}$ = $F^{1} \downarrow S$                   \Comment $S\times S$ downsampling
			\State $\NN^{2}$ = PatchMatch($F^{2},F^{0}$)            \Comment NN field at level 2
			\State $\NN^{1}_0$ = $\NN^{2} \uparrow S$               \Comment initial estimate of $\NN^{1}$
			\State $[M^{1},\NN^{1}]$ = CMD($F^{1},F^{0},\NN^{1}_0$) \Comment CMD at level 1
			\State $M^{0}_0$ = $M^{1} \uparrow S$                   \Comment $M^{0}_0$ gives the VoI
			\State $\NN^{0}_0$ = $\NN^{1} \uparrow S$               \Comment initial estimate of $\NN^{0}$
			\State $[M,\NN^{0}]$ = CMD($F^{0},\NN^{0}_0,$ VOI)      \Comment CMD at level 0 on VoI
		\end{algorithmic}
	\end{spacing}
	\caption{Multi-resolution Video Copy-Move Detector}
	\label{algo:MR-CMD}
\end{algorithm}

Therefore, the bulk of processing is now at level 1, where PatchMatch works in its standard configuration.
{\red
To further reduce the processing cost we resort again to $S \times S$ subsampling, and run PatchMatch at this level-2 resolution.
The retrieved NN field is then upsampled and used to initialize PatchMatch at level-1
in order to ensure its quick convergence, thus reducing the number of iterations.
Note that subsampling operates only to reduce the {\em source} features to match, while the target features are not sampled at all, otherwise the correct offset may not be found.
For example, at the lowest resolution, features drawn from $F^2$ are matched to features drawn from $F^0$.
Note that, thanks to the first-order predictors, propagation keeps working correctly.
In Fig.\ref{fig:Block_Diagram} we show a block diagram of the complete multiresolution scheme,
described in detail in the caption and, more formally, through the pseudo code of in Algorithm 1.

Finally, to gain some more speed, we resorted to parallel computing.
The parallel code works seamlessly for feature extraction and in the post-processing phase.
As for PatchMatch, each thread operates only on a portion of the source data (wile target data are not partitioned),
and the offset subfields are joined after each iteration.
By so doing, however, propagation of offsets across partition boundaries may be delayed significantly.
Therefore, we change partitions after each coupe of forward-backward iterations, orienting boundaries along columns, rows, or frames in round-robin fashion.
}

\begin{figure}[t]
\centering
\def\tbspc{2mm}
\begin{tabular}{c@{\hspace{\tbspc}}c@{\hspace{\tbspc}}c@{\hspace{\tbspc}}c@{\hspace{\tbspc}}c@{\hspace{\tbspc}}c}
\includegraphics[width=0.15\textwidth]{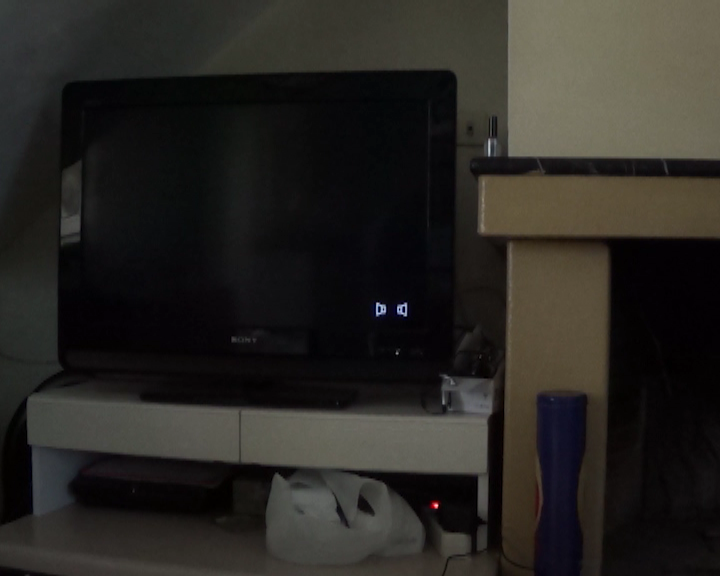}  &
\includegraphics[width=0.15\textwidth]{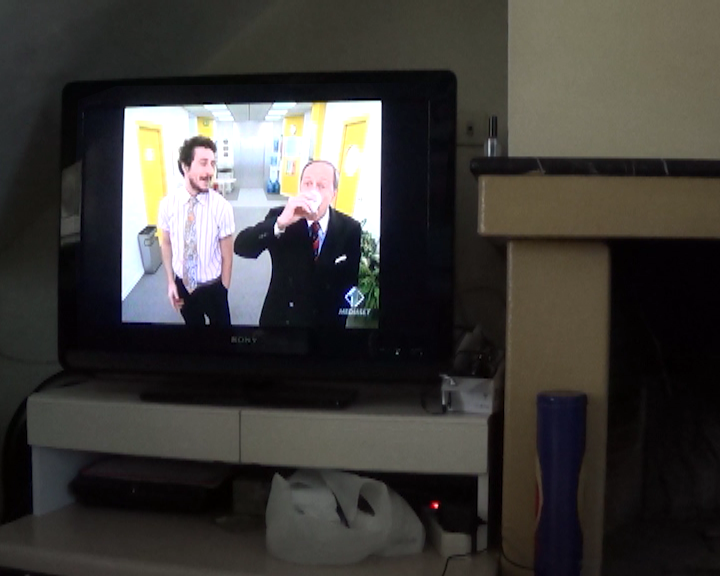} &
\includegraphics[width=0.18\textwidth]{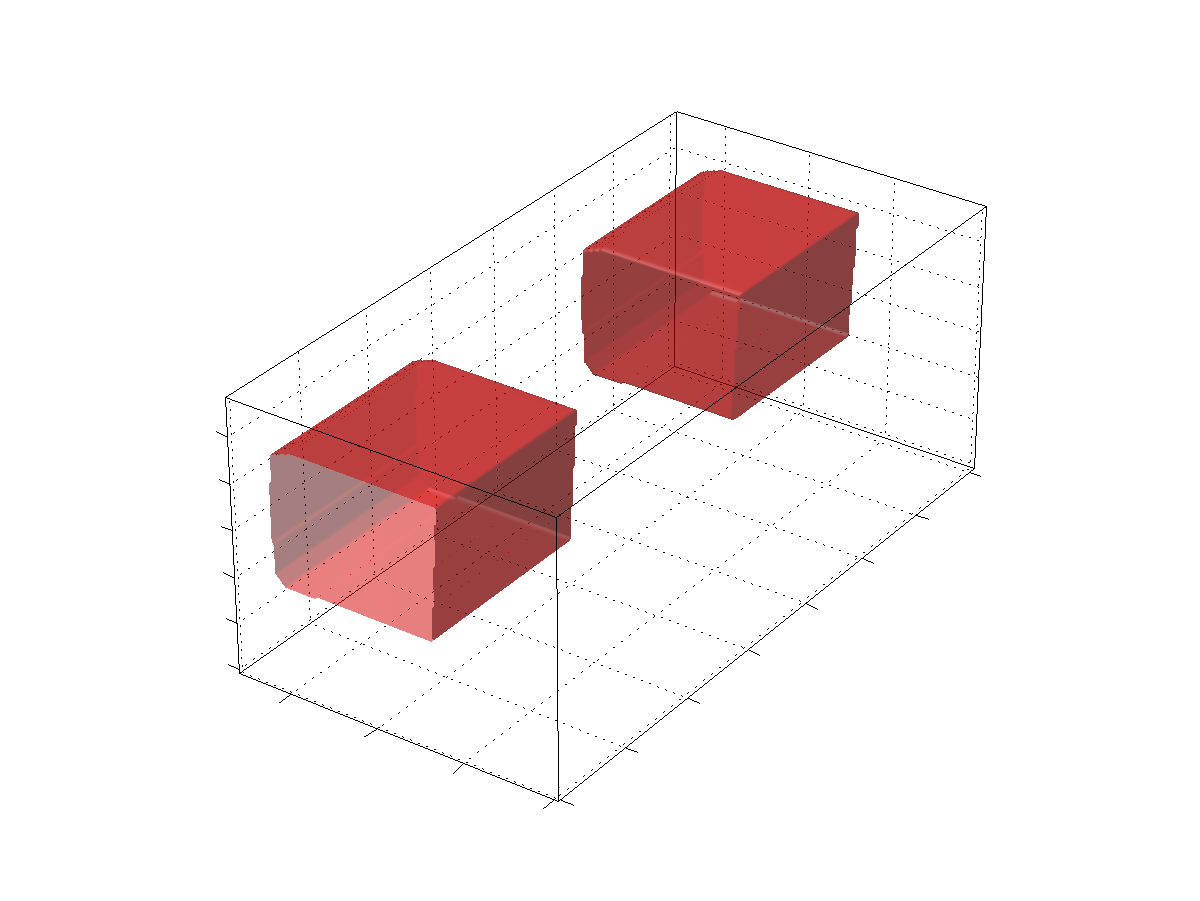}   \\
\multicolumn{3}{l}{\vspace{-4mm}} \\
\multicolumn{3}{l}{\footnotesize a) video \#1, TV screen: additive, large, static} \\
\multicolumn{3}{l}{\vspace{-10mm}} \\
\includegraphics[width=0.15\textwidth]{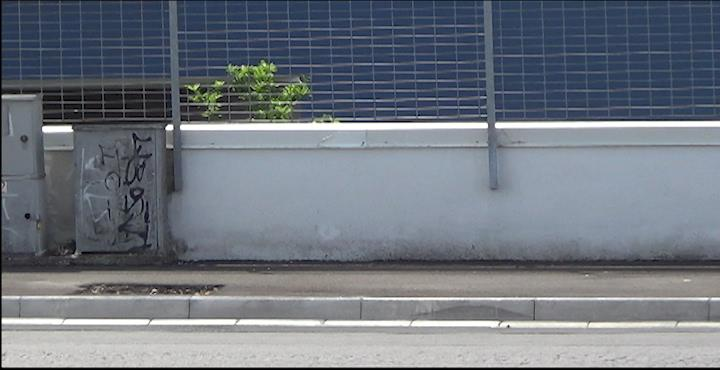}  &
\includegraphics[width=0.15\textwidth]{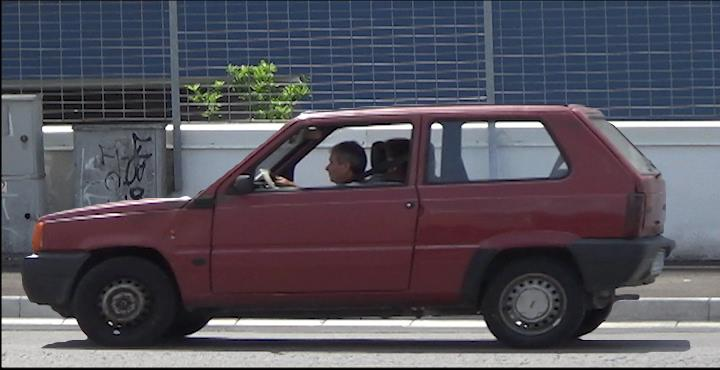} &
\includegraphics[width=0.18\textwidth]{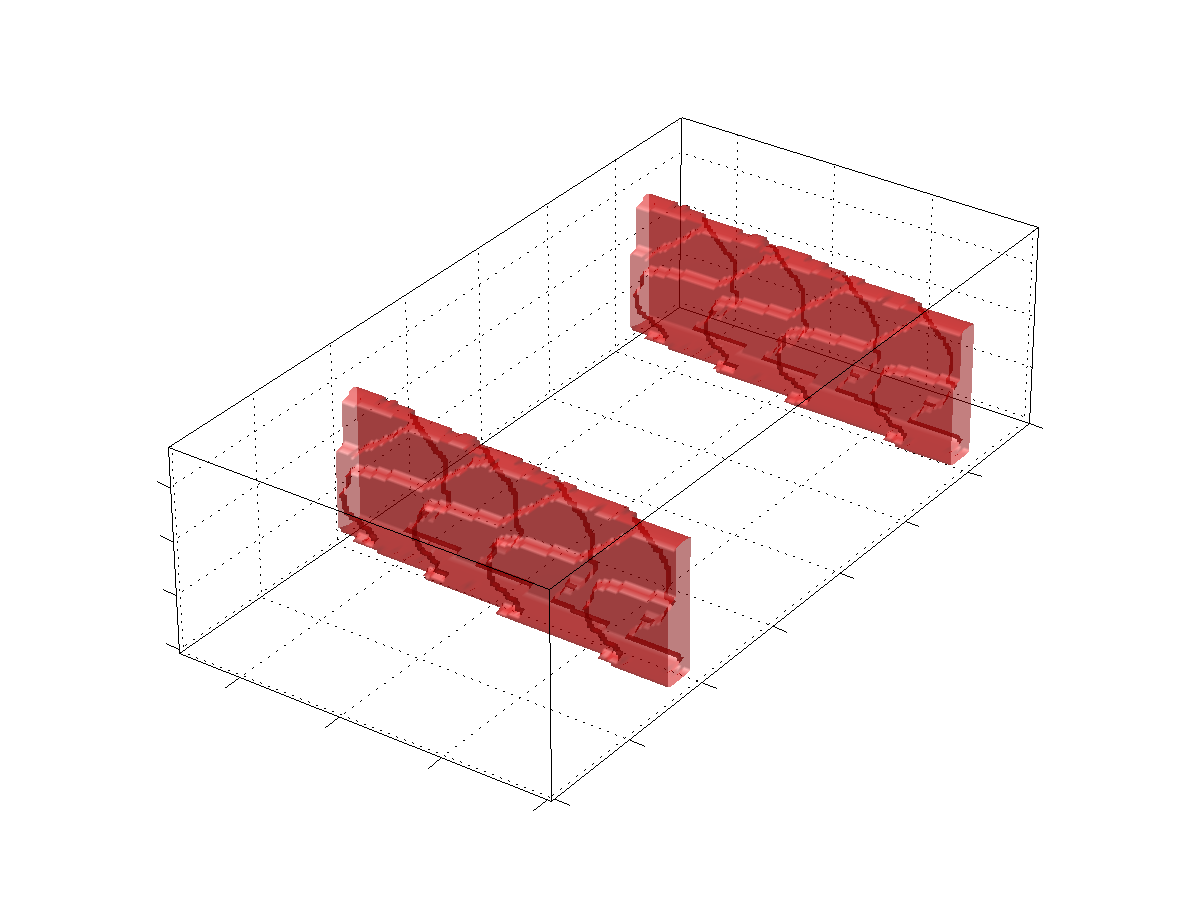}   \\
\multicolumn{3}{l}{\vspace{-4mm}} \\
\multicolumn{3}{l}{\footnotesize b) video \#2, Fast car: additive, large, fast (low depth)} \\
\multicolumn{3}{l}{\vspace{-6mm}} \\
\includegraphics[width=0.15\textwidth]{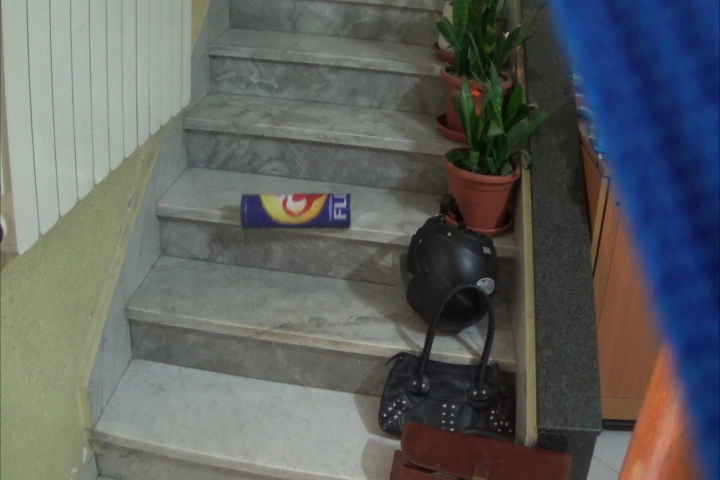}  &
\includegraphics[width=0.15\textwidth]{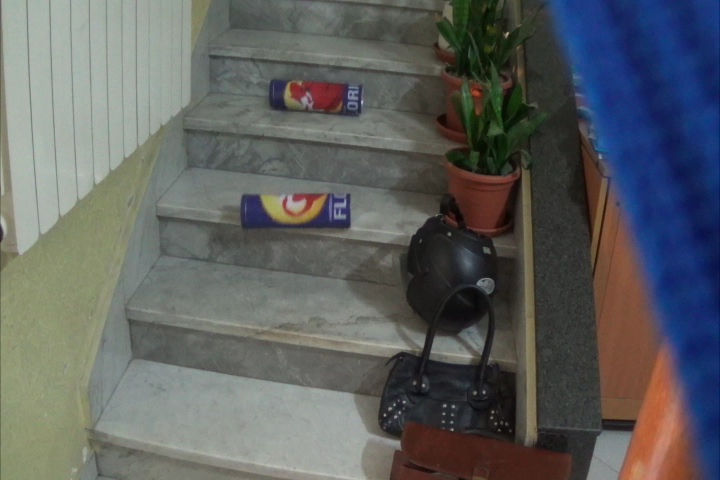} &
\includegraphics[width=0.18\textwidth]{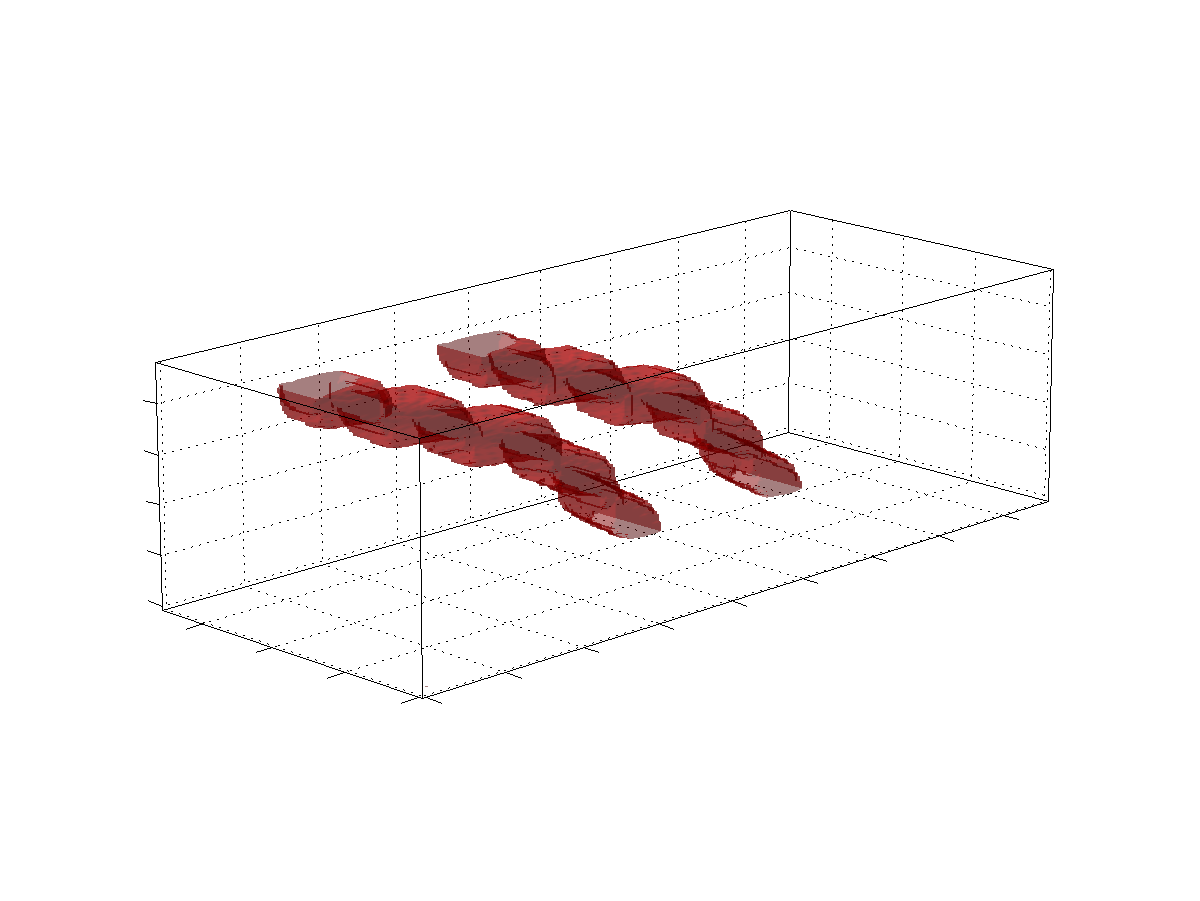}   \\
\multicolumn{3}{l}{\vspace{-4mm}} \\
\multicolumn{3}{l}{\footnotesize c) video \#5, Falling can: additive, small, fast (low depth)} \\
\multicolumn{3}{l}{\vspace{-6mm}} \\
\includegraphics[width=0.15\textwidth]{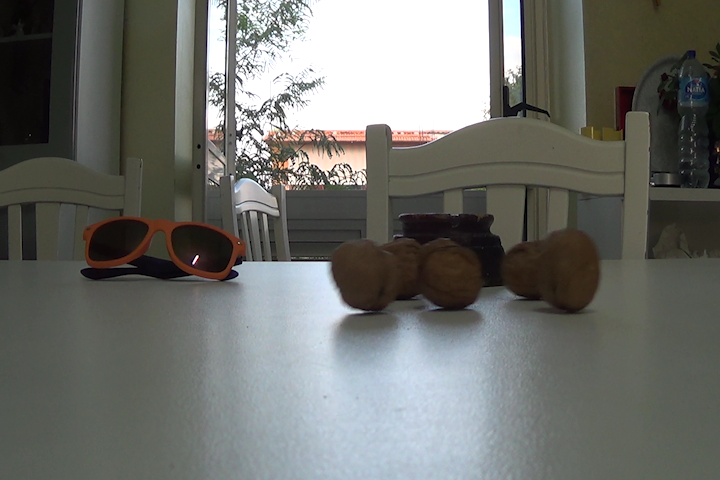}  &
\includegraphics[width=0.15\textwidth]{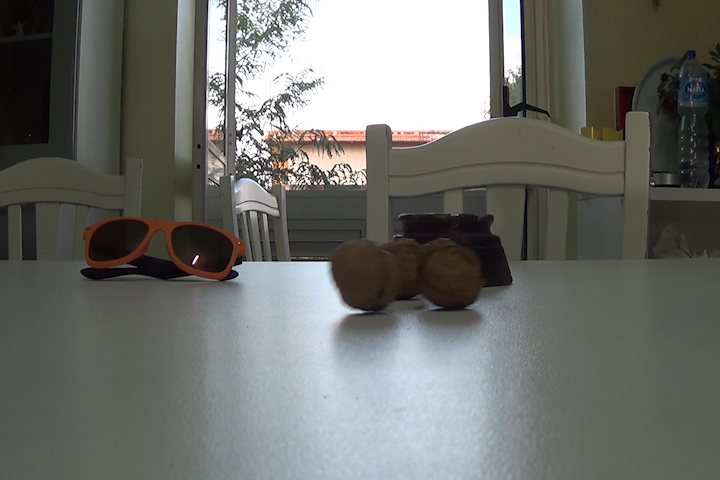} &
\includegraphics[width=0.18\textwidth]{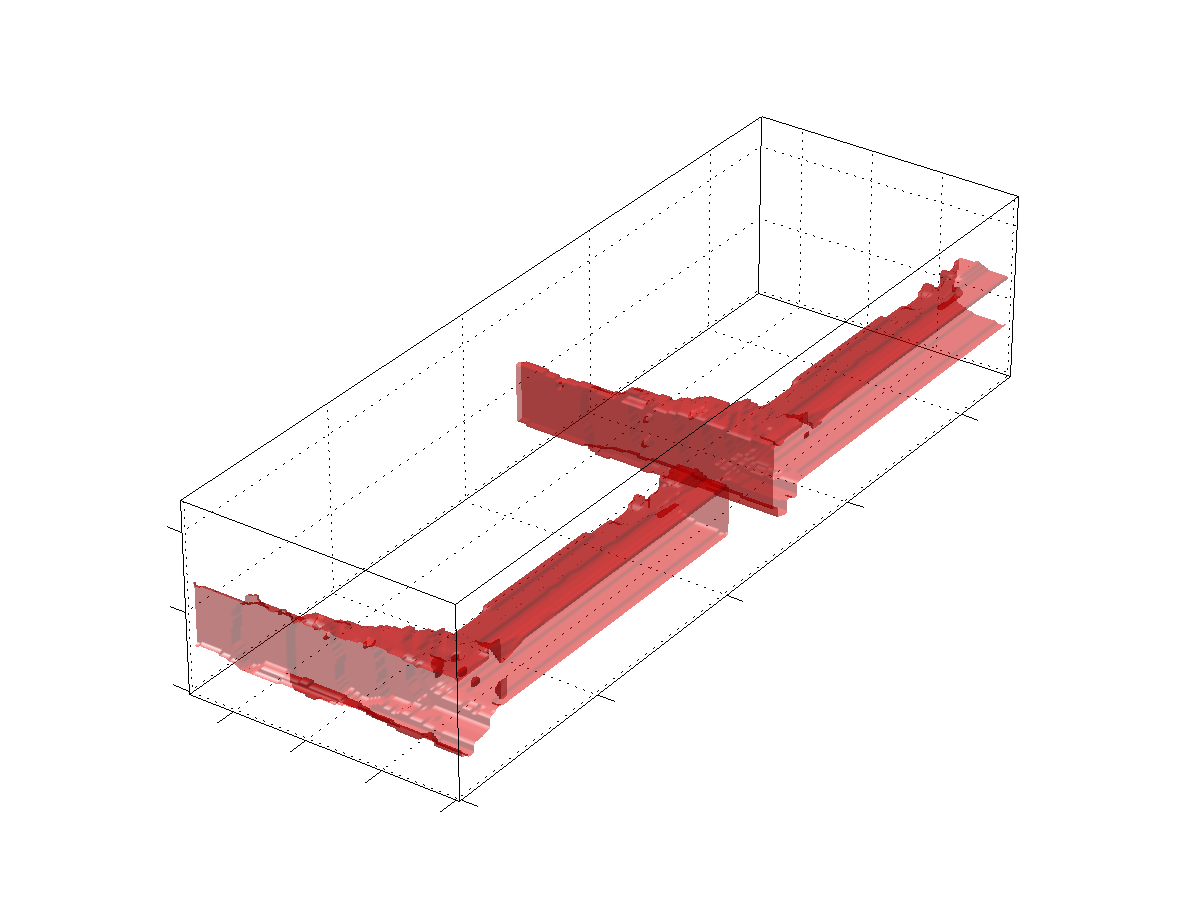}   \\
\multicolumn{3}{l}{\vspace{-4mm}} \\
\multicolumn{3}{l}{\footnotesize d) video \#6, Walnuts: occlusive, small, saturated area} \\
\multicolumn{3}{l}{\vspace{-0mm}} \\
\includegraphics[width=0.15\textwidth]{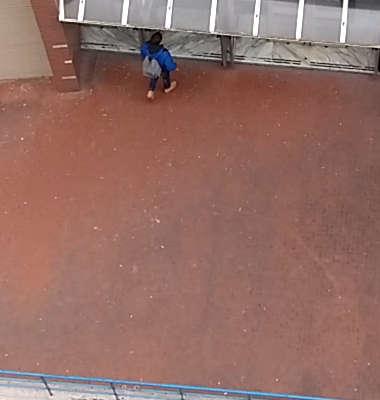}  &
\includegraphics[width=0.15\textwidth]{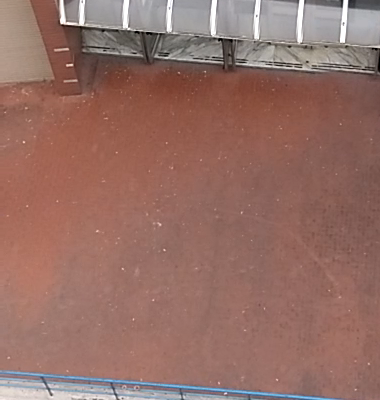} &
\includegraphics[width=0.18\textwidth]{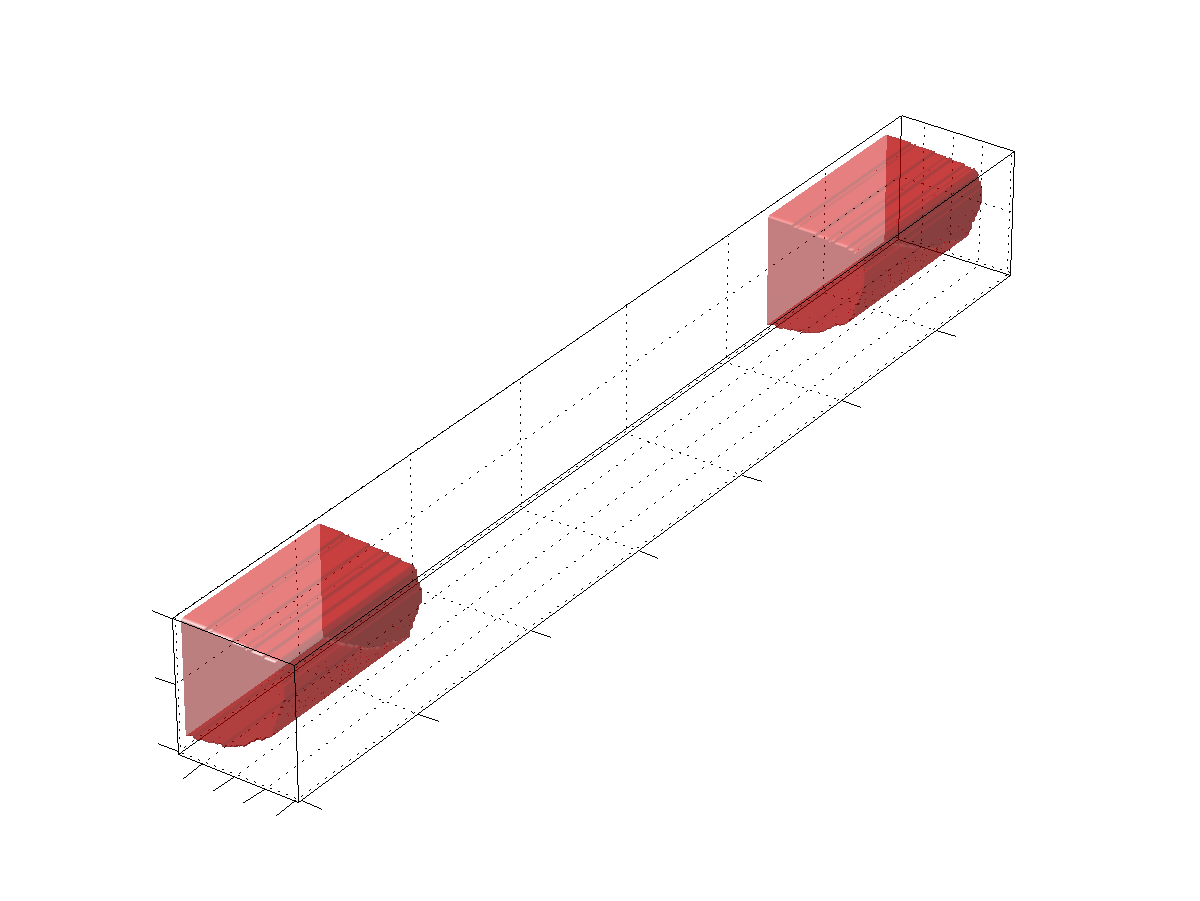}   \\
\multicolumn{3}{l}{\vspace{-4mm}} \\
\multicolumn{3}{l}{\footnotesize e) video \#11, Student: occlusive} \\
\multicolumn{3}{l}{\vspace{-0mm}} \\
\includegraphics[width=0.15\textwidth]{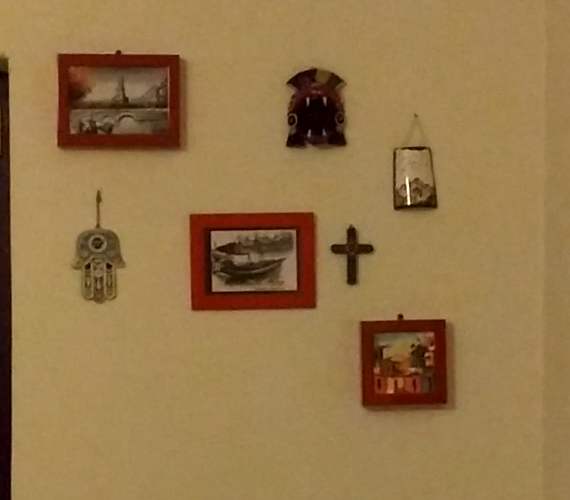}  &
\includegraphics[width=0.15\textwidth]{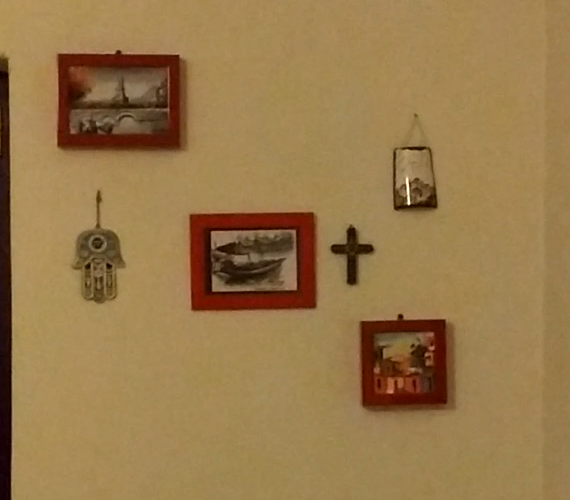} &
\includegraphics[width=0.18\textwidth]{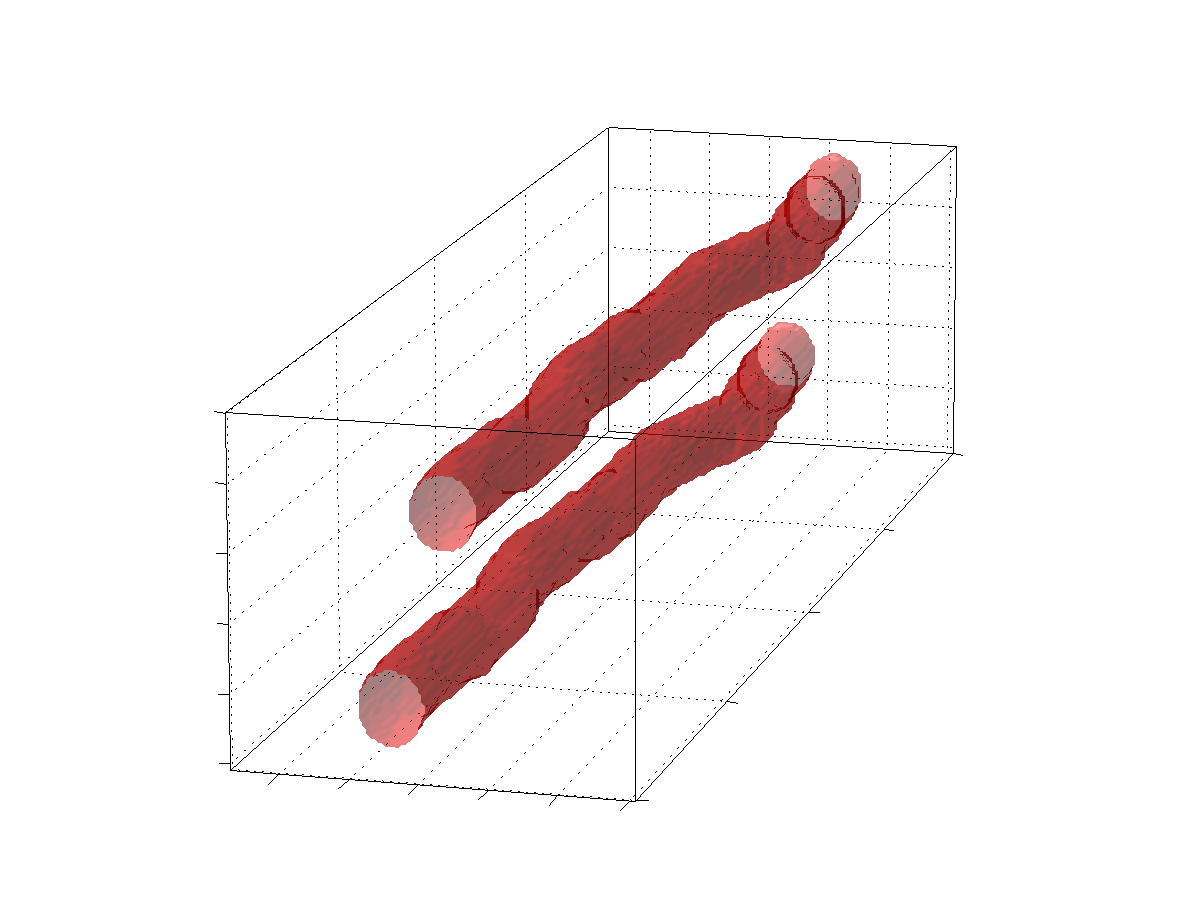}   \\
\multicolumn{3}{l}{\vspace{-4mm}} \\
\multicolumn{3}{l}{\footnotesize f) video \#14,  Wall frame: occlusive, small} \\
\multicolumn{3}{l}{\vspace{-2mm}} \\
\end{tabular}
\caption{Examples from the GRIP dataset.
From left to right: original frame, copy-moved frame, 3D view of the ground truth.
In the 3D views, the origin of the axes is the bottommost point, and the time axis is on bottom-right.}
\label{fig:GRIP_Videos}
\end{figure}

\section{Experimental analysis}

The performance of the proposed method was assessed through a number of experiments under various operative conditions.
In the following we will describe the datasets, the performance measures, and finally the experimental results and comparisons.

\renewcommand{\ru}{\rule{0mm}{4mm}}
\renewcommand{\tabcolsep}{3pt}
\begin{table}
\footnotesize
\centering
\begin{tabular}{|r|l|c|r||c|r|r|c|c|} \hline	
\multicolumn{4}{|c||}{\ru video} & \multicolumn{5}{c|}{copy-move} \\ \hline
\ru \# & name         &     frame size & frames & add./occ. & $\rho_{\max}$ & $d_{\max}$ &  rot. & flp. \\ \hline\hline
\ru  1 & TV screen    & 576$\times$720 &  141~~ &       add &        182.7~ &       43~  &  \chk & \chk \\ \hline
\ru  2 & Fast Car     & 370$\times$720 &  140~~ &       add &        203.2~ &        9~  &       & \chk \\ \hline
\ru  3 & Felt-Tip Pen & 550$\times$720 &  100~~ &       add &         62.3~ &        4~  &  \chk & \chk \\ \hline
\ru  4 & Rolling Can  & 480$\times$660 &  125~~ &       add &        229.6~ &       18~  &  \chk & \chk \\ \hline
\ru  5 & Falling Can  & 480$\times$720 &  174~~ &       add &         71.3~ &       29~  &       &      \\ \hline
\ru  6 & Walnuts      & 480$\times$720 &  221~~ & occlusive &        199.5~ &      102~  &       &      \\ \hline
\ru  7 & Can 1        & 520$\times$720 &  201~~ &       add &        220.6~ &       28~  &       & \chk \\ \hline
\ru  8 & Can 2        & 720$\times$720 &  210~~ &       add &        112.1~ &       15~  &  \chk & \chk \\ \hline
\ru  9 & Lamp         & 390$\times$465 &  455~~ &       add &        159.9~ &      129~  &  \chk &      \\ \hline
\ru 10 & Tennis Ball  & 640$\times$360 &  200~~ &       add &        195.4~ &       31~  &       & \chk \\ \hline
\ru 11 & Student      & 400$\times$380 &  340~~ & occlusive &        176.3~ &       60~  &       &      \\ \hline
\ru 12 & Cell 1       & 400$\times$500 &   92~~ &       add &         63.7~ &       92~  &  \chk & \chk \\ \hline
\ru 13 & Cell 2       & 512$\times$512 &   92~~ & occlusive &        107.5~ &       92~  &  \chk & \chk \\ \hline
\ru 14 & Wall Frame   & 500$\times$570 &  200~~ & occlusive &         50.6~ &      155~  &  \chk &      \\ \hline
\ru 15 & Statue       & 590$\times$480 &  100~~ & occlusive &         65.9~ &       61~  &       &      \\ \hline
\end{tabular}
\caption{Features of the GRIP dataset}
\label{tab:GRIP_dataset}
\end{table}

\subsection{Dataset}

We prepared a dataset, called the GRIP dataset from now on, comprising 15 short videos with rigid copy-moves, 10 additive and 5 occlusive.
They were carried out by the first author using After Effects Pro, a tool for video editing.
As a result, there are little or no artifacts which may raise suspects on the video, just as would happen with a real-world skilled attacker.
Nonetheless, since we consider rather short videos,
additive copy-moves may be obvious anyway, since the same object appears twice in a few seconds.
On the contrary, it seems safe to say that occlusive copy-moves can be hardly spotted without specific tools.
In addition, by using rotation or temporal flipping, when meaningful, also additive copy-moves become less visible.
All copy-moved videos will be available online at http://www.grip.unina.it/ together with their pristine versions and the ground truths.

\renewcommand{\ru}{\rule{0mm}{4mm}}
\newcommand{\Smu}{{$\Sigma,\mu$}}
\renewcommand{\tabcolsep}{5pt}
\begin{table*}
	\footnotesize
	\centering
	\begin{tabular}{|c||c|c|c|r||c|c|c|r||c|c|c|r||c|c|c|r||c|c|c|r|} \hline
		\ru       & \multicolumn{4}{c||}{Basic 2D} & \multicolumn{4}{c||}{Basic 3D} & \multicolumn{4}{c||}{Fast 2D} & \multicolumn{4}{c||}{Fast 3D} & \multicolumn{4}{c|}{Bestagini}  \\ \hline
		\ru video &    det. & f.a. & ~~F~~ & time  &   det.  & f.a. & ~~F~~ & time  &  det. &  f.a. & ~~F~~ &  time &   det. & f.a. & ~~F~~ &  time &        det. & f.a. & ~F~ & time \\ \hline\hline
		\ru     1 &    \chk &      & 0.96  & 15.42 &   \chk  &      & 0.95  & 17.50 & \chk  &       &  0.97 &  2.17 &   \chk &      &  0.97 &  3.25 &        \chk &      &  -- &  8.9 \\ \hline
		\ru     2 &    \chk &      & 0.88  & 15.45 &   \chk  &      & 0.68  & 17.19 & \chk  &       &  0.78 &  2.77 &   \chk &      &  0.67 &  3.44 &        \chk &      &  -- &  7.3 \\ \hline
		\ru     3 &    \chk &      & 0.56  & 16.39 &   \chk  &      & 0.29  & 23.24 & \chk  &       &  0.60 &  2.67 &   \chk &      &  0.31 &  3.00 &        \chk &      &  -- &  6.7 \\ \hline
		\ru     4 &    \chk &      & 0.88  & 14.92 &   \chk  &      & 0.79  & 16.75 & \chk  &       &  0.88 &  2.77 &   \chk &      &  0.76 &  3.32 &        \chk &      &  -- &  7.2 \\ \hline
		\ru     5 &    \chk &      & 0.84  & 16.70 &   \chk  &      & 0.86  & 20.29 & \chk  &       &  0.81 &  2.07 &   \chk &      &  0.86 &  3.24 &        \chk & \chk &  -- & 14.9 \\ \hline
		\ru     6 &    \chk & \chk & 0.72  & 16.50 &   \chk  &      & 0.74  & 18.58 & \chk  &  \chk &  0.73 &  2.35 &   \chk &      &  0.81 &  3.45 &             &      &  -- & 11.7 \\ \hline
		\ru     7 &    \chk &      & 0.83  & 18.45 &   \chk  &      & 0.78  & 20.25 & \chk  &       &  0.90 &  2.54 &   \chk &      &  0.81 &  3.41 &        \chk & \chk &  -- & 11.5 \\ \hline
		\ru     8 &    \chk &      & 0.87  & 19.73 &   \chk  &      & 0.77  & 24.23 & \chk  &       &  0.89 &  2.20 &   \chk &      &  0.76 &  3.32 &        \chk & \chk &  -- & 15.2 \\ \hline
		\ru     9 &    \chk &      & 0.93  & 17.80 &   \chk  &      & 0.92  & 20.31 & \chk  &       &  0.94 &  2.40 &   \chk &      &  0.93 &  4.02 &        \chk & \chk &  -- & 14.4 \\ \hline
		\ru    10 &    \chk &      & 0.91  & 15.69 &   \chk  &      & 0.89  & 16.67 & \chk  &       &  0.94 &  2.30 &   \chk &      &  0.92 &  3.45 &        \chk & \chk &  -- &  6.3 \\ \hline
		\ru    11 &    \chk & \chk & 0.88  & 14.14 &   \chk  &      & 0.87  & 18.00 & \chk  &  \chk &  0.86 &  3.05 &   \chk &      &  0.88 &  4.15 &             &      &  -- &  7.7 \\ \hline
		\ru    12 &    \chk &      & 0.80  & 16.23 &   \chk  &      & 0.77  & 18.78 & \chk  &       &  0.87 &  1.96 &   \chk &      &  0.83 &  3.81 &             &      &  -- &  2.6 \\ \hline
		\ru    13 &    \chk &      & 0.91  & 15.43 &   \chk  &      & 0.90  & 18.26 & \chk  &       &  0.92 &  2.49 &   \chk &      &  0.91 &  4.02 &             &      &  -- &  4.4 \\ \hline
		\ru    14 &    \chk &      & 0.74  & 16.66 &   \chk  &      & 0.71  & 19.42 & \chk  &       &  0.77 &  2.35 &   \chk &      &  0.77 &  3.39 &             &      &  -- &  8.8 \\ \hline
		\ru    15 &    \chk &      & 0.72  & 16.05 &   \chk  & \chk & 0.51  & 20.17 &       &  \chk &  0.00 &  2.26 &   \chk & \chk &  0.41 &  3.32 &             &      &  -- &  3.8 \\ \hline\hline
		\ru  \Smu &     15  &    2 & 0.83  & 16.37 &     15  &    1 & 0.76  & 19.31 &  14   &  3    &  0.79 &  2.42 &   15   &  1   &  0.75 &  3.51 &           9 &    5 &     &  8.8 \\ \hline
	\end{tabular}
	\caption{Detection, localization and efficiency performance for plain copy-moves on the GRIP dataset}
	\label{tab:GRIP_plain_full}
\end{table*}

Tab.\ref{tab:GRIP_dataset} shows synthetic statistics of all videos and forgeries.
Moreover, for some selected videos,
Fig.\ref{fig:GRIP_Videos} shows a sample frame of the original and forged video,
together with a 3D view of the ground truth which provides some immediate insight on the spatio-temporal structure of the forgery.
Large and static copy-moves, like that of Fig.\ref{fig:GRIP_Videos}(a) will be easily detected in any condition.
On the contrary, small and fast-moving copy-moves represent a severe challenge.
Note that by ``fast'', we mean a copy-move with a rapidly changing mask,
maybe spanning just a few frames at any pixel, like in both Fig.\ref{fig:GRIP_Videos}(b) and Fig.\ref{fig:GRIP_Videos}(c),
while the speed of the physical object inside the mask is immaterial for our aim.
To capture synthetically these geometric features we use the max-radius and max-depth indicators.
The max-radius is defined as $\rho_{\max}=\max_t \sqrt{A(t)/\pi}$, with $A(t)$ being the area of the tampered region in frame $t$.
Likewise, max-depth is $d_{\max}=\max_s d(s)$, with $d(s)$ the depth of the tampered region for pixel $s$, possibly much smaller than the total copy-move duration.
In Fig.\ref{fig:GRIP_Videos}(d)-(f) we show some occlusive forgeries.
The first one may be especially challenging, giving rise to a large number of false alarms due to the saturated area in the middle.

Besides our own dataset,
we also consider the REWIND dataset, described in \cite{Bestagini2013} and available online\footnote{https://sites.google.com/site/rewindpolimi/downloads/datasets/video-copy-move-forgeries-dataset}.
This dataset, however, comprises only rigid additive copy-moves and comes without a ground truth.
In addition, some videos (e.g., Duck, Fast Car) appear to be splicings rather then copy moves
(maybe with material taken from part of the video not available to the user)
or else the copied regions have been subjected to some unreported processing before pasting them back.
For these reasons, we use REWIND only for some experiments, turning to the GRIP dataset for a more reliable analysis.

\subsection{Performance evaluation}

\begin{figure}[t]
\centering
\begin{minipage}[c]{.48\linewidth} \centerline{\epsfig{figure=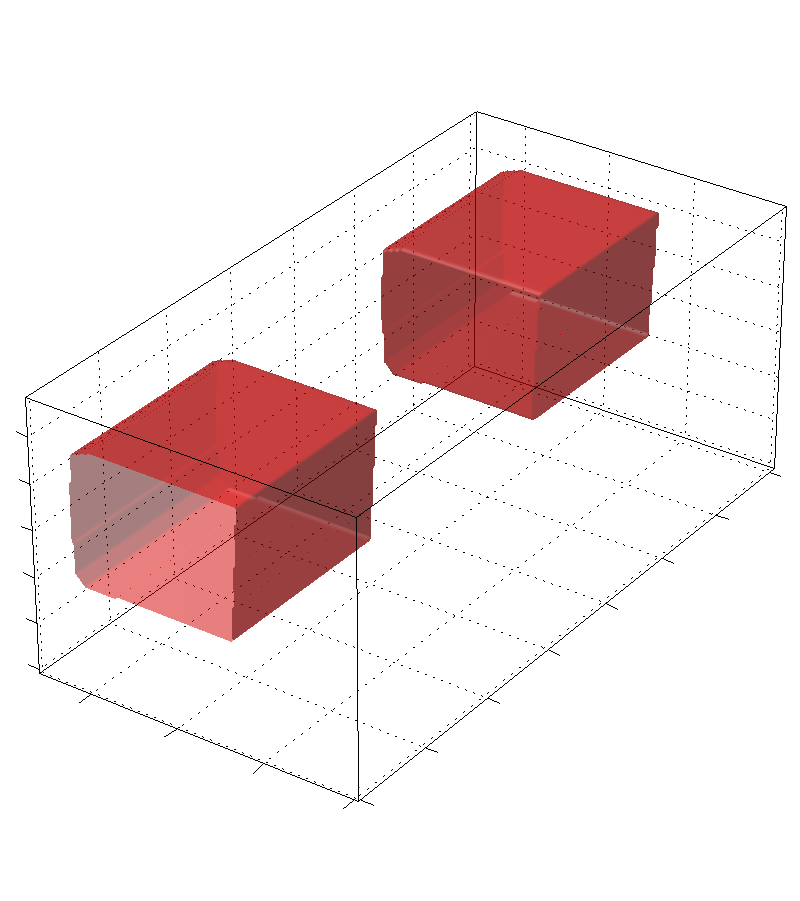, width=4.0cm}} \end{minipage} \hfill
\begin{minipage}[c]{.48\linewidth} \centerline{\epsfig{figure=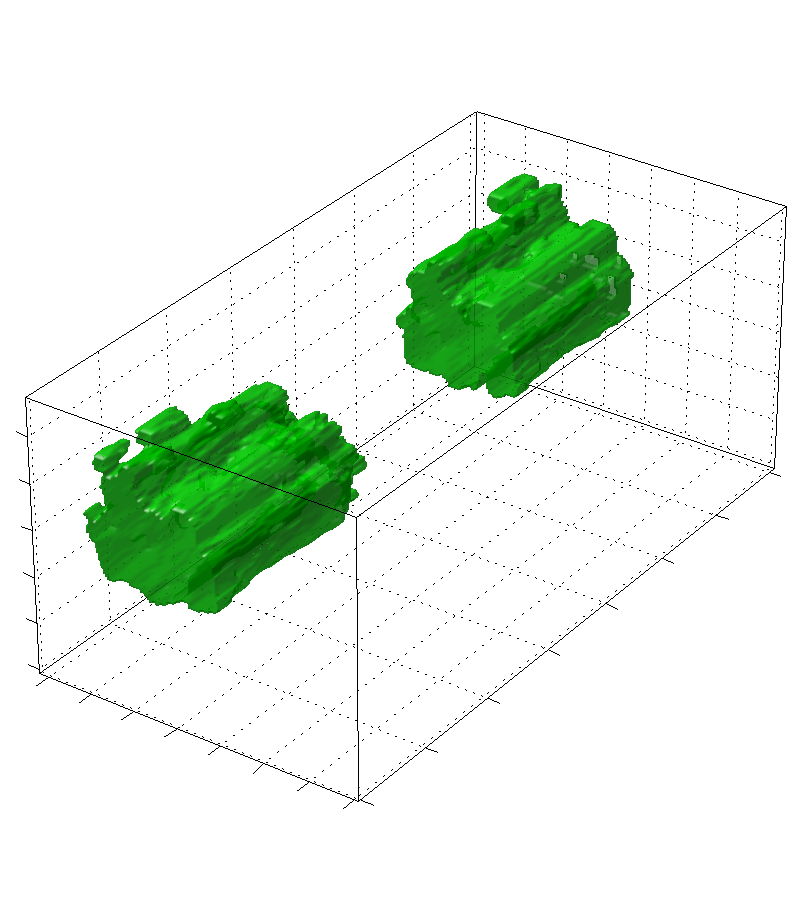,  width=4.0cm}} \end{minipage} \hfill
\caption{Examples of ground truth ($GT$) and detection map ($M$).
In $GT$ (left) copy-moved regions, both original and clones, are set to 1 (red).
In $M$ (right), detected copy-moves are set to 1 (green).
}
\label{fig:Example_Measures}
\end{figure}

{\red
The performance is measured in terms of detection and localization accuracy, besides processing time.
Detection is declared if, after the post-processing, which includes the removal of small regions, a large number of detected pixels are present in the output map $M$
\begin{equation}
    |M| > T_{\rm detection}
\end{equation}
where $|x|$ counts the number of ones in $x$ and
the threshold $T_{\rm detection}$ is set for each version of the algorithm by preliminary experiments.
When a copy-move is actually present, this is a correct detection, otherwise it is a false alarm.
Therefore, to quantify false alarms, in the experiments we run our detectors also to the original videos.
}

\renewcommand{\ru}{\rule{0mm}{4mm}}
\renewcommand{\tabcolsep}{5pt}
\begin{table*}
\footnotesize
\centering
\begin{tabular}{|c|c|c||c|c|c||c|c|c||c|c|c||c|c|c||c|c|c|} \hline
\multicolumn{3}{|c||}{\ru}           &            \multicolumn{3}{c||}{Basic 2D} & \multicolumn{3}{c||}{Basic 3D} & \multicolumn{3}{c||}{Fast 2D} & \multicolumn{3}{c||}{Fast 3D} & \multicolumn{3}{c|}{Bestagini}  \\  \hline
              dataset & \ru case     &          \# videos  & det. & f.a. & ~~F~~ &            det. & f.a. & ~~F~~ &           det. & f.a. & ~~F~~ &           det. & f.a. & ~~F~~ &               det. & f.a. & ~F~ \\ \hline\hline
                GRIP  & \ru plain    &                 15  &   15 &    2 &  0.83 &              15 &    1 &  0.76 &             14 &    3 & 0.79  &             15 &    1 &  0.75 &                 ~9 &    5 &  -- \\ \hline\hline
\multirow{4}{*}{GRIP} & \ru QF = 10  & \multirow{4}{*}{15} &   15 &    1 &  0.84 &              15 &    1 &  0.77 &             14 &    2 & 0.74  &             14 &    1 &  0.75 &                 ~9 &    5 &  -- \\ \cline{2-2}\cline{4-18}
                      & \ru QF = 15  &                     &   15 &    1 &  0.76 &              15 &    1 &  0.72 &             13 &    2 & 0.65  &             15 &    1 &  0.70 &                 ~9 &    4 &  -- \\ \cline{2-2}\cline{4-18}
                      & \ru QF = 20  &                     &   12 &    1 &  0.54 &              12 &    1 &  0.56 &             13 &    2 & 0.53  &             12 &    0 &  0.52 &                 ~9 &    5 &  -- \\ \hline\hline
\multirow{4}{*}{GRIP} & \ru \ag{ 5}  & \multirow{4}{*}{ 8} &   ~8 &   -- &  0.81 &              ~7 &   -- &  0.73 &             ~5 &   -- & 0.40  &             ~7 &   -- &  0.68 &                 ~2 &   -- &  -- \\ \cline{2-2}\cline{4-18}
                      & \ru \ag{25}  &                     &   ~7 &   -- &  0.71 &              ~4 &   -- &  0.60 &             ~3 &   -- & 0.25  &             ~4 &   -- &  0.44 &                 ~2 &   -- &  -- \\ \cline{2-2}\cline{4-18}
                      & \ru \ag{45}  &                     &   ~5 &   -- &  0.56 &              ~4 &   -- &  0.43 &             ~2 &   -- & 0.12  &             ~4 &   -- &  0.43 &                 ~2 &   -- &  -- \\ \hline\hline
                GRIP  & \ru flipping &                  9  &   ~8 &   -- &  0.81 &              ~9 &   -- &  0.76 &             ~6 &   -- & 0.59  &             ~7 &   -- &  0.59 &                 ~3 &   -- &  -- \\ \hline\hline
              REWIND  & \ru plain    &                 10  &   ~8 &    4 &    -- &              ~9 &    4 &    -- &             ~8 &    4 &   --  &             ~6 &    1 &    -- &                 ~6 &    3 &  -- \\ \hline
	\end{tabular}
	\caption{Summary of detection and localization performance on the whole set of experiments}
	\label{tab:ALL_synthetic}
\end{table*}

{\red
To measure the localization performance we define the sets
\begin{itemize}
\item   TP (true positive): detected copy-move pixels;
\item   FP (false positive): detected pristine pixels;
\item   TN (true negative): undetected pristine pixels;
\item   FN (false negative): missed copy-move pixels.
\end{itemize}
(see also Fig.\ref{fig:Detection_Maps}) from which the F-measure indicator is derived as
\begin{equation}
    F = \frac{2|{\rm TP}|}{2|{\rm TP}|+|{\rm FP}|+|{\rm FN}|}
\end{equation}
If detection map and ground truth coincide, then $|{\rm FP}| = |{\rm FN}| = 0 $, and the F-measure reaches its maximum value, equal to 1.
However, as the number of false negative or false positive pixels increase, the F-measure decreases rapidly.
In particular, the F-measure is more informative than the overall accuracy when the two classes of interest are very unbalanced,
which is the case of typical forged videos, where only a small fraction of the data are tampered with.
}


Finally, we measure efficiency in terms of normalized CPU-time, s/Mpixel,
that is, the time required to process the whole video divided by its size in Mpixel.
CPU-times refer to a computer with a 2GHz Intel Xeon processor with 16 cores, 64GB RAM and GPU Nvidia GeForce GTX Titan X.

\subsection{Numerical results}

In Tab.\ref{tab:GRIP_plain_full} we report results for the GRIP dataset in the presence of plain copy-moves,
involving only rigid spatio-temporal translations, and possibly some local processing at the boundary of the copied area to reduce artifacts.
No rotation and flipping are allowed, here, and no global post-processing, such as compression, and noise addition.
For each technique and each video
we mark with a \chk\ symbol whether the copy move is detected (det),
and whether a false alarm (f.a.) is declared, namely a copy-move is detected in the pristine video where there are none.
Then we report the F-measure, to quantify localization accuracy, and the normalized CPU-time.
{\red
We use features of the length 12, in the 2D case, and length 18, in the 3D case,
the latter obtained by considering 6 Zernike moments over 3 consecutive frames and computing the even-odd transform.
In the fast versions, a subsampling step $S=4$ is used.
}

To provide some comparison with the state-of-the-art, we include also the technique proposed by Bestagini {\it et al.} \cite{Bestagini2013} which, however, addresses only the detection task.
Unfortunately, literature techniques conceived for localization \cite{Hsu2008, Subramanyam2012}
make very restrictive hypotheses on the forged videos, hence they cannot be used on realistic datasets as GRIP or REWIND.
We have also excluded from comparison the 3D version of PatchMatch working on RGB values \cite{Bleyer2011, Newson2013},
since it can detect only rigid copy-moves, as also shown in \cite{Damiano2015},
and does not provide any advantage over the feature-based version.

{\red
Performance figures are very good for all variants of the proposed algorithm.
The basic version of the algorithm, without multi-resolution processing, detects all copy-moves,
both with single-frame 2D features (Basic-2D algorithm, from now on) and with 3D flip-invariant features (Basic-3D), with very few false alarms.
On the other hand, a few false alarms in this context are not really critical
since they raise attention on some candidate copy-moves that may be analyzed with much greater care afterwards.
On the contrary, missed detection cannot be recovered easily.
The fast versions of the algorithms (Fast-2D and Fast-3D) are also quite reliable.
Only Fast-2D misses one copy-move, of the occlusive type, probably because of the loss of spatial synchronization due to subsampling.
The reference method \cite{Bestagini2013}, instead, misses all occlusive copy-moves and also an additive one,
besides originating a slightly larger number of false alarms.

The localization performance is extremely high in all cases.
Barring video 15, missed by Fast-2D, the only critical case seems to be video 3, and only for the Basic-3D and Fast-3D algorithms.
This is easily explained by noting that $d_{\max}=4$, for this video, namely, the copy-move is extremely thin in time,
causing inaccuracies at the temporal boundaries when 3D features are used.
Nonetheless, these problems do not prevent correct detection.

\begin{figure*}[t]
\centering
\includegraphics[width=0.48\textwidth]{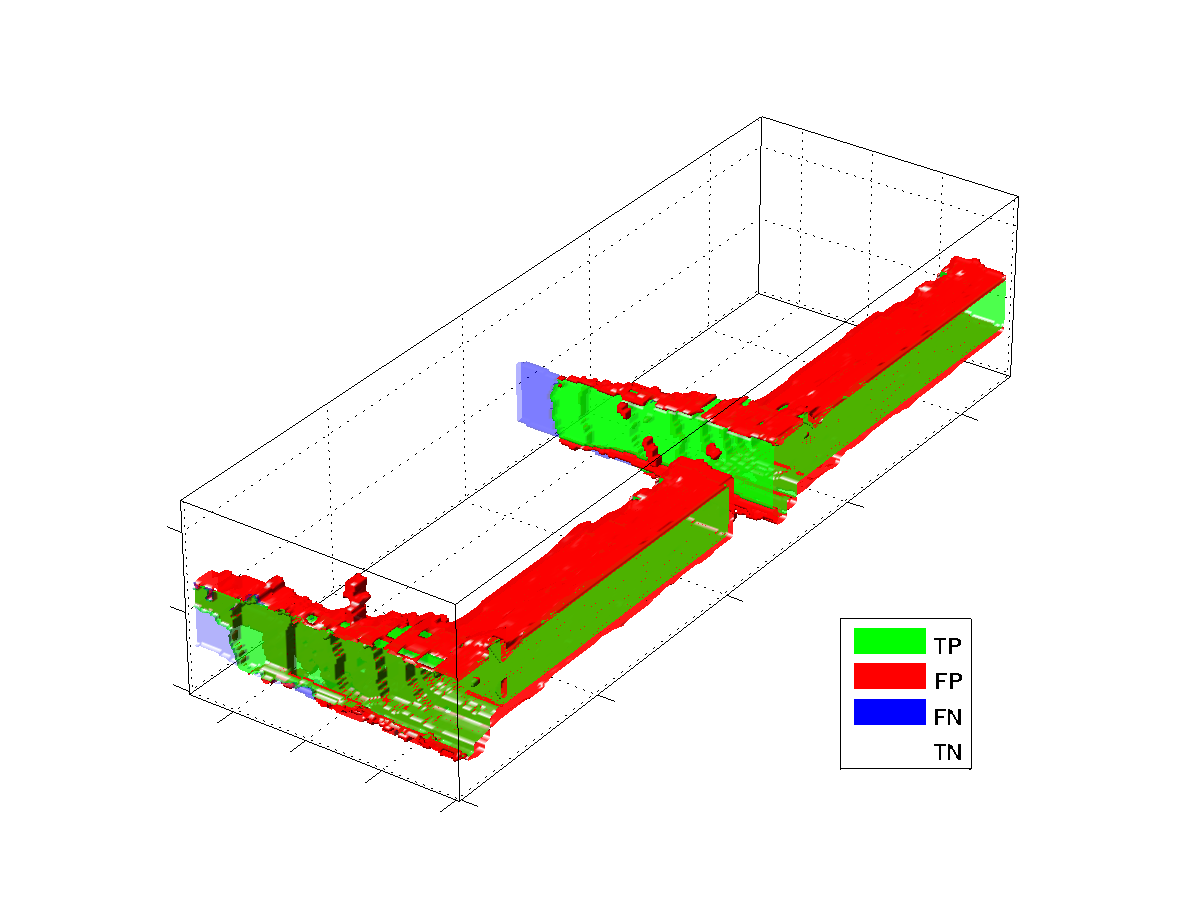}
\includegraphics[width=0.48\textwidth]{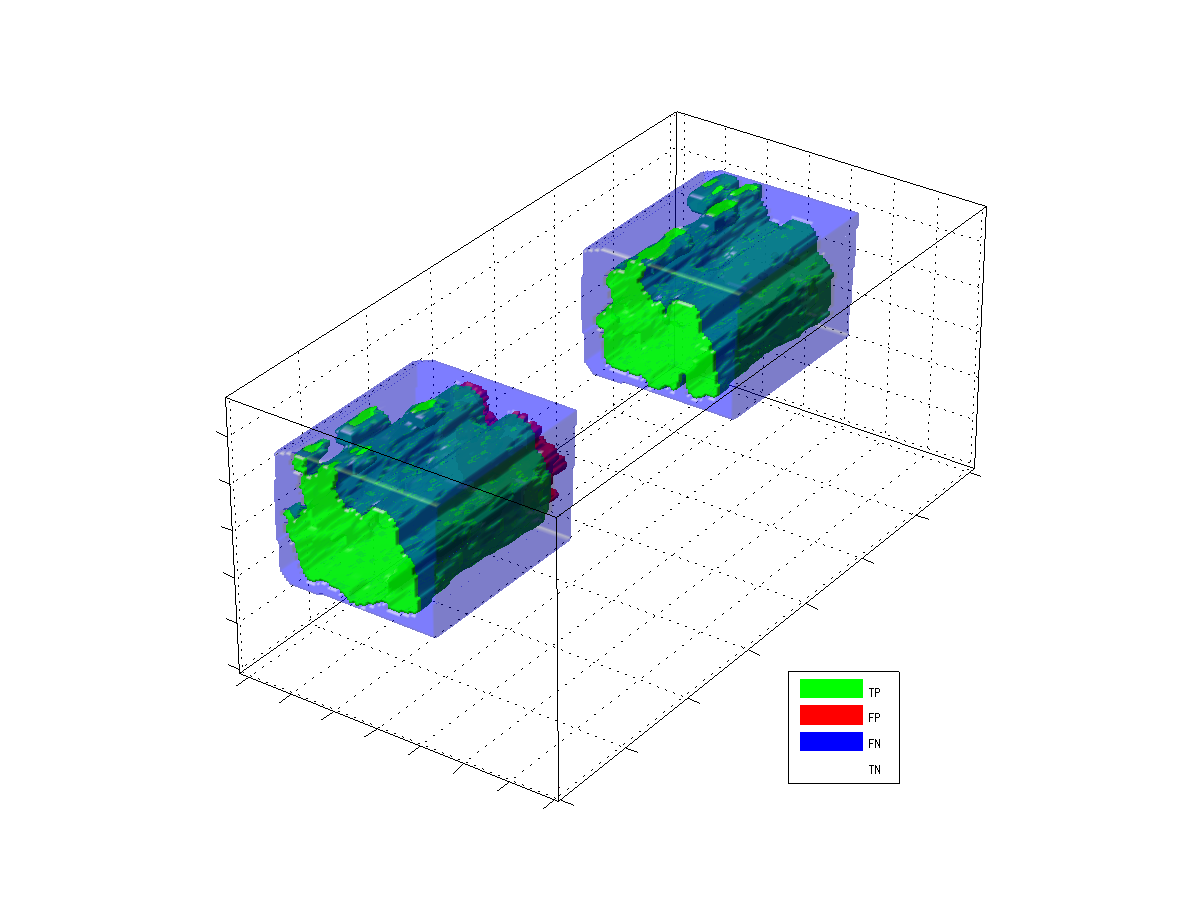} \\
{\footnotesize Plain copy-move in video \#6 Walnuts             \hspace{24mm} Copy-move with compression at QF=20 in video \#1 TV screen} \\
\includegraphics[width=0.48\textwidth]{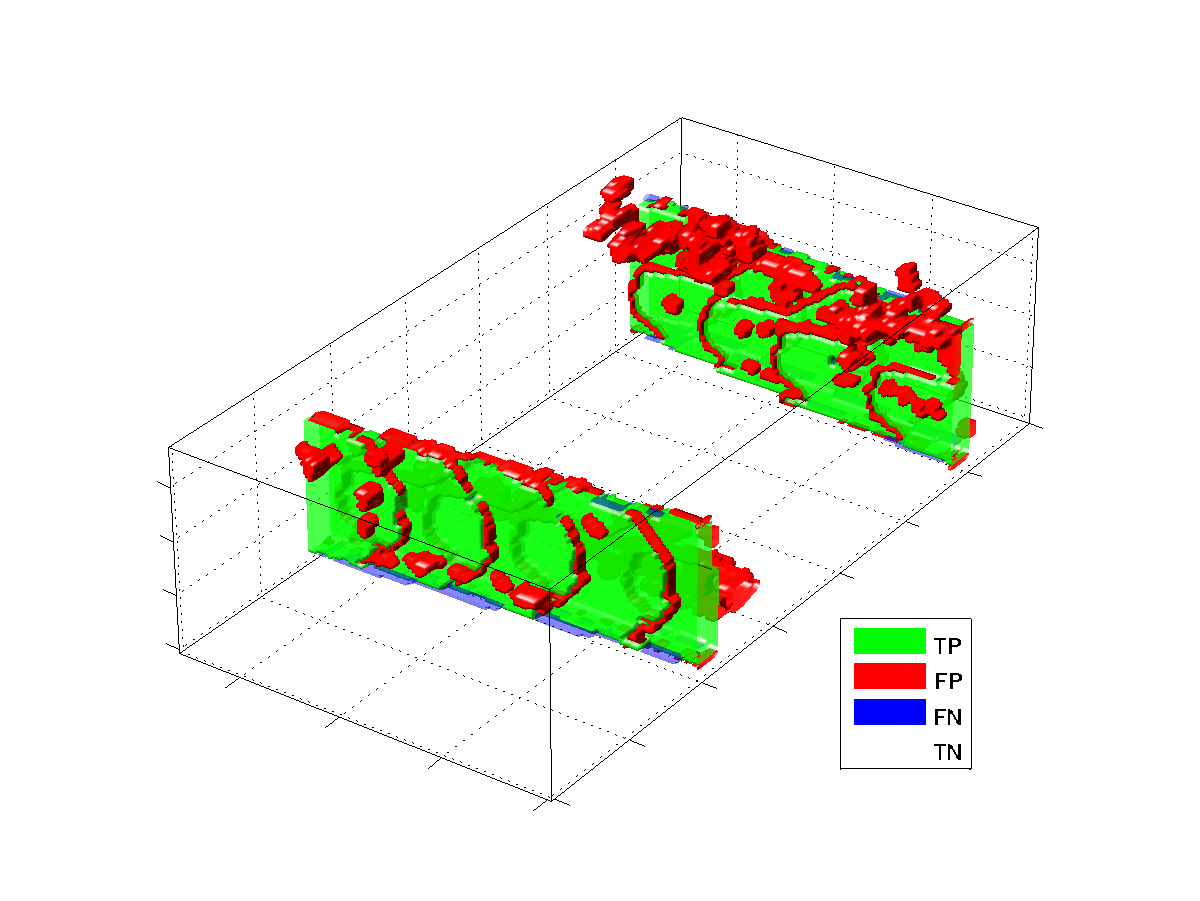}
\includegraphics[width=0.48\textwidth]{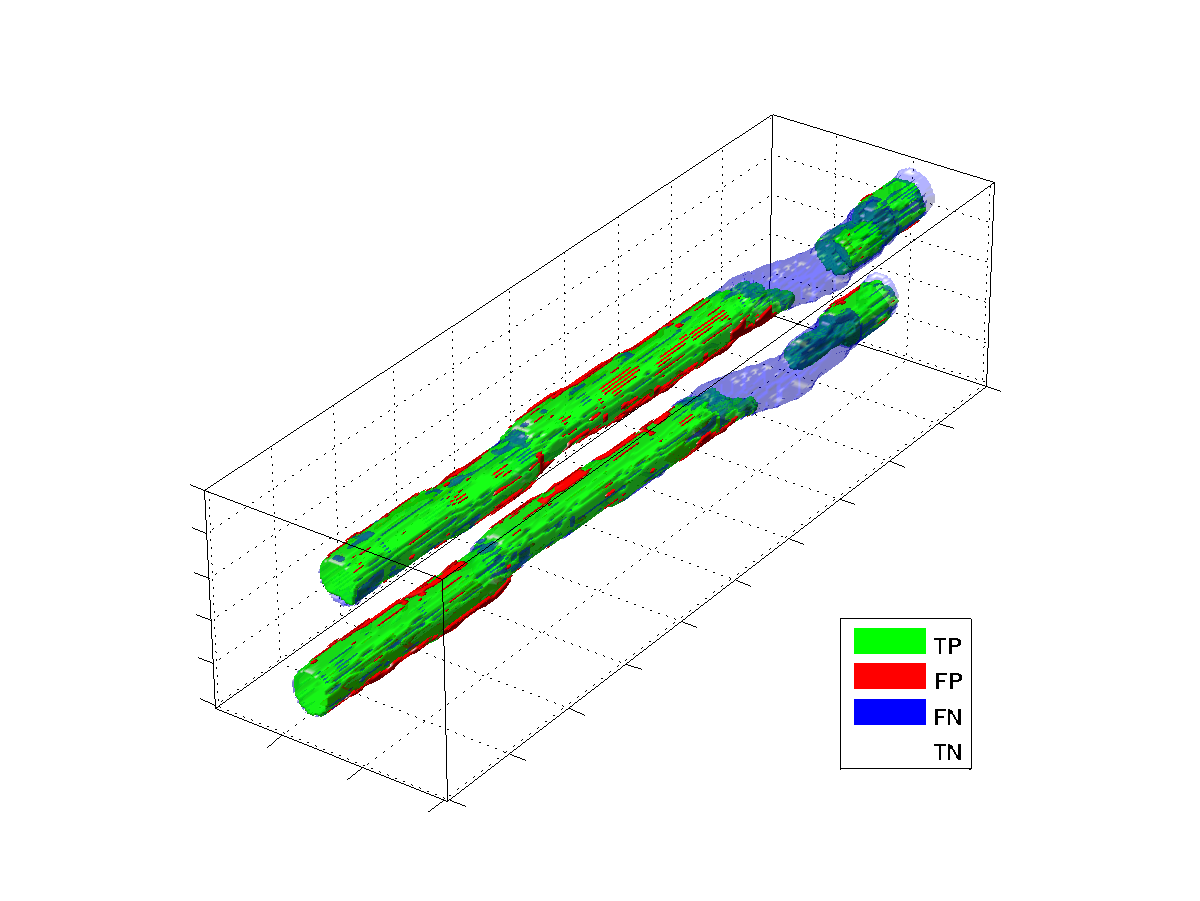} \\
{\footnotesize Copy-move with flipping in video \#2 Fast Car    \hspace{24mm} Copy-move with 45$^o$ rotation in video \#14 Wall Frame}
\caption{Sample color-coded detection maps for videos of the GRIP dataset. TN pixels are transparent for correct visualization.
The plain (top-left) and flipped (bottom-left) copy moves are fully detected, with only some inaccuracies at the boundaries.
Compression (top-right) impacts on localization performance, especially on the saturated areas of the video object.
Rotation (bottom-right) causes the loss of a copy moved segment, following sudden camera motion.}
\label{fig:Detection_Maps}
\end{figure*}

Turning to processing speed,
the overall CPU times scale pretty rigidly with the video size (frame size $\times$ number of frames).
In fact, the Basic 2D and 3D algorithms take about 16.4 s/Mpixel and 19.3 s/Mpixel, respectively, with very small deviations across the videos.
The fast versions are indeed much faster,
bringing the average CPU-time down to 2.9 and 3.5 s/Mpixel, respectively, the difference mainly due to the longer 3D features.
}

Let us now analyze performance in more challenging situations,
namely, in the presence of video compression, and of copy-move rotation and flipping.
Experimental results are reported in Tab.\ref{tab:ALL_synthetic}, only in synthetic form for the sake of brevity.
In the same table we also report results obtained in the absence of further processing on the GRIP dataset (GRIP plain, first line)
and on the REWIND dataset (REWIND plain, last line).

Compressed videos are more the norm than the exception, and studying performance in this situation is of paramount importance.
To this end, we consider MPEG compression at quality factors QF = 10, 15, and 20, which correspond roughly to high, medium and low quality sources.
Together with the forged videos, we compress also the pristine ones, which allows us to compute both detection and false alarm figures.
{\red
The basic version of the algorithm keeps providing an excellent performance with both 2D and 3D flip-invariant features,
although some missed detections are observed in the most challenging case of QF = 20.
A similar behavior is observed with the fast versions, with just a few further missed detections.
The F-measure remains always quite large (lower values are mostly due to missed detections), indicating a very good localization ability.
In all cases, a very small number of false alarms is observed.
Note that the reference technique detects only 9 copy-moved video, with a larger number of false alarms and,
as already said, does not have localization ability.

In lines 5-7 of Tab.\ref{tab:ALL_synthetic} we analyze the case of video objects that are rotated before pasting,
which may be due to composition needs (small angles) or made on purpose to fool copy-move detectors.
The analysis applies only to the 8 videos with rotated copy-moves,
while no false alarm analysis is possible, since objects are not rotated in the original videos.
The basic algorithm keeps working very well at small angles,
while at large angles, 25 or 45 degrees, the version with 3D features exhibits a large number of missed detections.
This is likely due to temporal boundary discontinuities, quite relevant for videos with small $d_{\max}$.
On the other hand, 3D features seem more robust when moving to the fast version, with no further missed detection,
while the Fast-2D algorithm exhibits a limited reliability.
The reference algorithm, instead, looks definitely unreliable with rotated copy-moves at all angles.

In the presence of flipping,
the proposed algorithm works very well with 3D flip-invariant features, with no missed detection for the basic version and 2 for the fast version,
slightly better than when 2D features are used.
Together with previous results,
this suggests using 3D flip-invariant features with the fast algorithm, while 2D features seem slightly preferable with the basic algorithm.
However,
more data are necessary to draw solid conclusions.}

Finally, let us consider the REWIND dataset.
In this case, no algorithm is able to provide perfect detection.
The best result is obtained with Basic-3D, but even in this case there is one missed detection and 4 false alarms.
However,
this is to be ascribed to the video themselves
since some forgeries appear to be splicings rather than copy-moves (notably the fast car video), which fully justifies the failures.
Indeed, the reference algorithm, tested by the authors on this very same dataset, provides an even poorer performance.

We conclude this subsection by showing, in Fig.\ref{fig:Detection_Maps}
some sample detection maps obtained with the proposed algorithm (basic, 2D features) on GRIP videos with copy-moves and various operating conditions.
The performance is always very good, although a whole segment is missed in the rotated copy-move, due to the large rotation angle,
highlighting the challenges raised by post-processing for detection.

\subsection{Complexity}

\begin{figure}	
\includegraphics[scale=0.25]{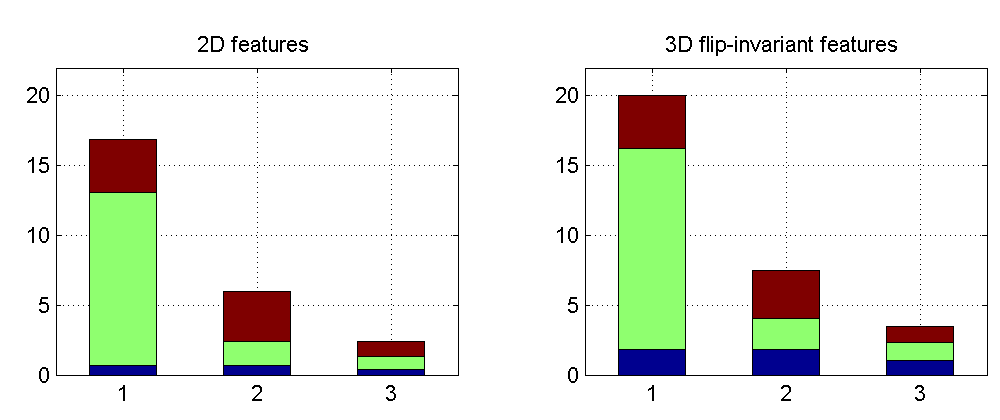}
\caption{Computational cost of feature extraction (blue), matching (green), and post-processing (red) phases
for various versions of the proposed algorithm using 2D features (left) or 3D flip-invariant features (right).
The dominant source of complexity in the basic algorithm (bar 1) is matching.
Multiresolution processing (bar 2) reduces sharply this cost.
Parallel computing (bar 3) further reduces the cost of all phases.
The final speed-up w.r.t. the basic version is about 7 with 2D features and 6 with 3D flip-invariant features.}
\label{fig:CPU}
\end{figure}

{\red
As already said,
a major effort has been devoted in this work to devise a computationally efficient technique.
The bar graphs of Fig.\ref{fig:CPU} describe the results for the case of 2D features (left) and 3D flip-invariant features (right)
in terms of normalized CPU times (s/Mpixel) averaged over all experiments, comprising a grandtotal of 153 videos.
From left to right,
the bars refer to the basic algorithm (1), its multi-resolution version (2), and the parallel implementation of the latter (3),
while colors identify feature extraction (blue), matching (green) and post-processing (red).
The multi-resolution processing impacts only on the matching phase,
largely reducing its cost and bringing total CPU-time
from 16.85 to 5.99 s/Mpixel with 2D features and
from 20.06 to 3.47 s/Mpixel with 3D features.
The parallel implementation, instead, reduces the cost of all phases, although to different degrees,
bringing the total CPU-time to 2.43 and 3.47 s/Mpixel, respectively.
Overall, Fast-2D guarantees a 7$\times$ speed-up w.r.t. Basic-2D, and Fast-3D a 6$\times$ speed-up w.r.t. Basic-3D,
with differences due mainly to the longer features used in the second case.

Still, even this large improvement cannot solve all computational problems of video copy-move detection.
To process a 1-minute video at 25 frames/s, with 0.5 Mpixel frames, about 30 minutes of CPU time are necessary.
Hence, a mass screening is probably out of reach.
However, one is often interested in analyzing only a specific video, or part of it, and willing to spend the necessary resources to obtain reliable information.
This is certainly the case of videos presented as evidence in court, where the fragments of interest are typically quite short.

On the other hand,
the complexity of copy-move detection is inherently quadratic with the size of the source,
since, in principle, all features must be compared with one another.
For the small videos of the GRIP dataset, one should compute in the order of $10^7$ distances per feature.
Thanks to PatchMatch, our basic algorithms reduce this number to about $10^2$.
Our fast parallel version is 6-7 times faster than that.
This timing may still be filed a little more, but not orders of magnitude.
Interestingly, the method proposed in \cite{Bestagini2013}, based on Fourier-domain analysis, is much slower than Fast-2D and Fast-3D
and the same applies to a simple 3D version of the keypoint-based method proposed in \cite{Amerini2011}.
}

\subsection{A real-world case: the Varoufakis video}

\begin{figure*}[t]
	\centering
	\begin{minipage}[c]{.32\linewidth} \centerline{\includegraphics[width=5.5 cm]{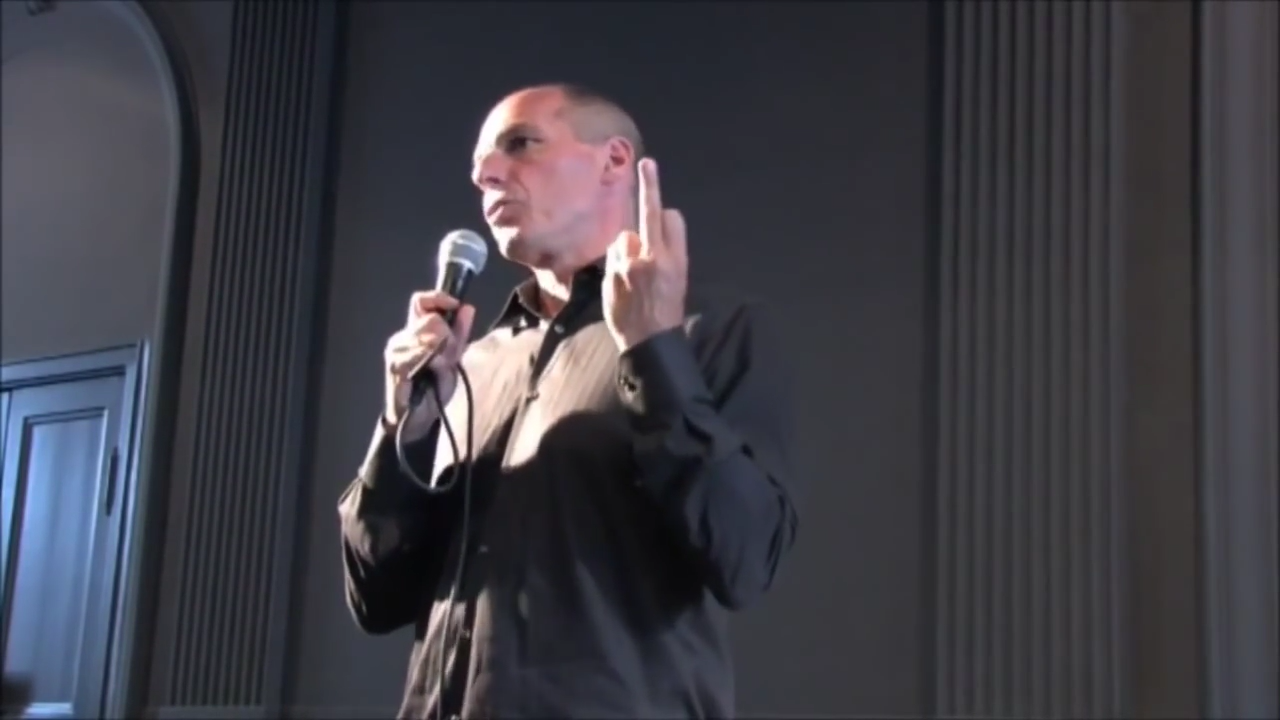}}   \end{minipage} \hfill
	\begin{minipage}[c]{.32\linewidth} \centerline{\includegraphics[width=5.5 cm]{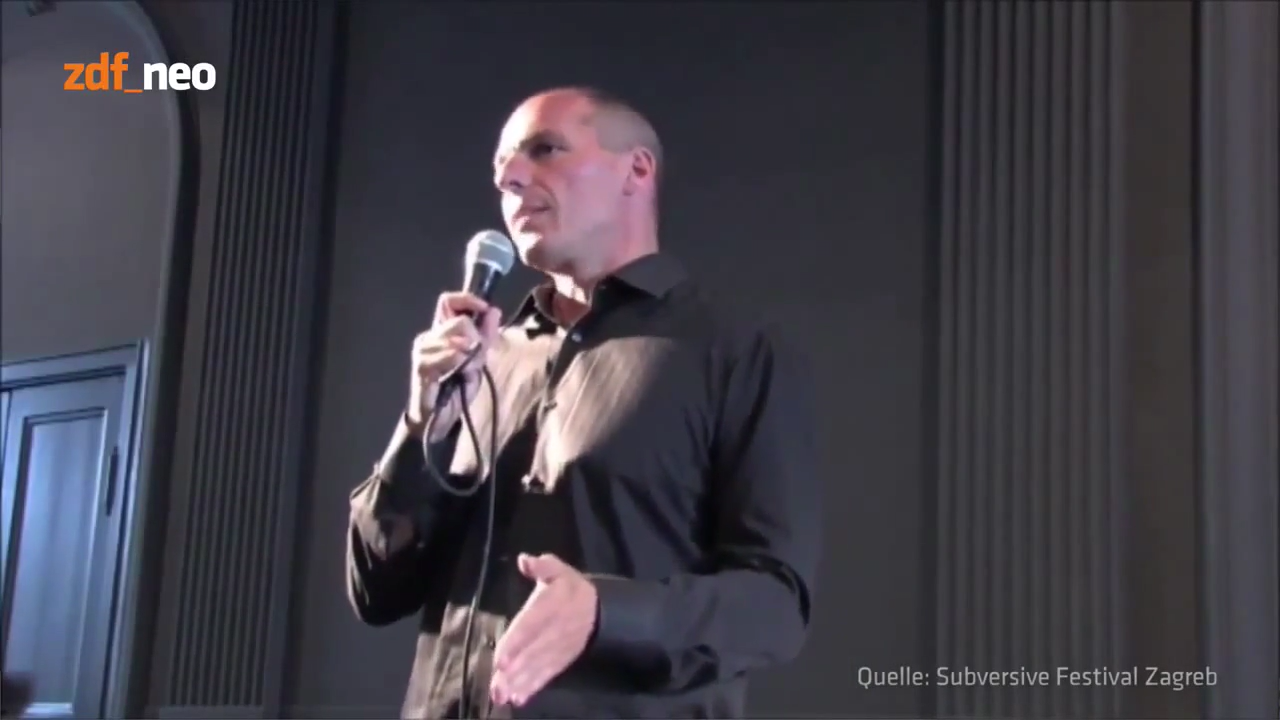}} \end{minipage} \hfill
	\begin{minipage}[c]{.32\linewidth} \centerline{\includegraphics[width=5.5 cm]{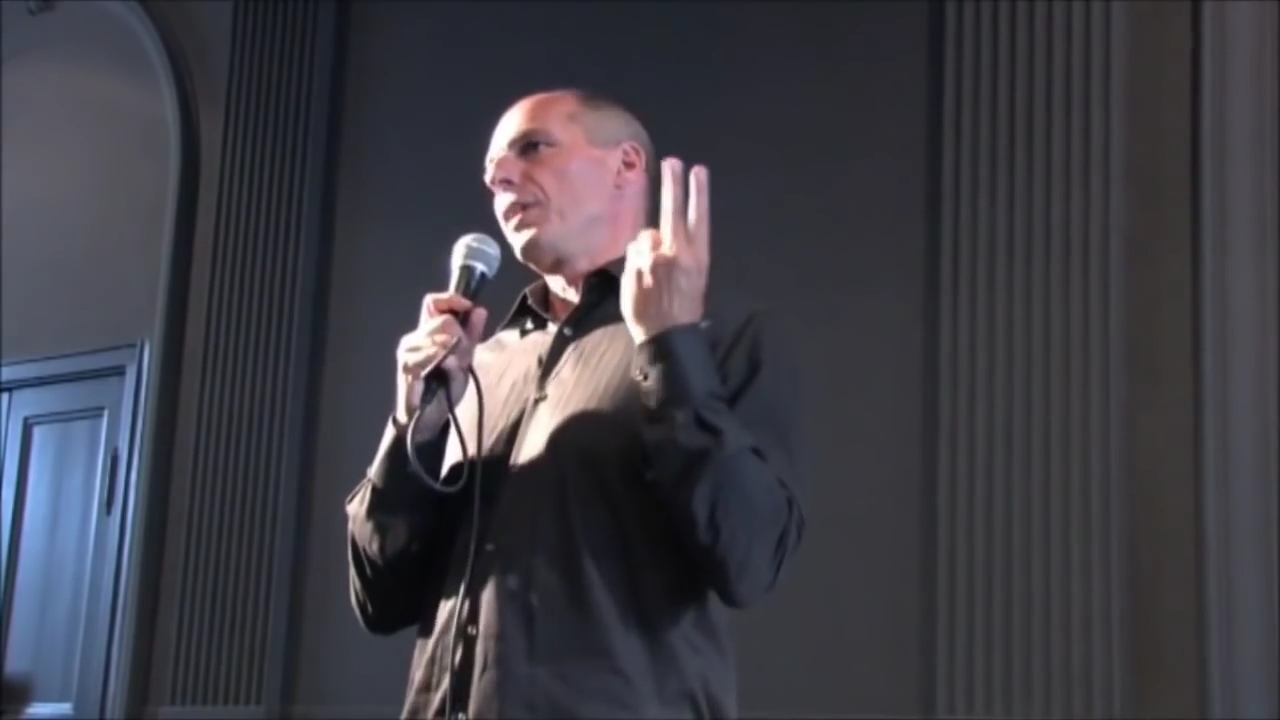}} \end{minipage} \hfill
	\caption{Frames taken from the three Varoufakis videos. From left to right: \#1, sticking middle finger, \#2, arm down, \#3, victory sign.}
\label{fig:Varoufakis_Frames}
\end{figure*}

We tested our algorithm on a real-world case that recently made the headlines all over the world, the well-known Varoufakis video.
While politicians of the European Union (EU) were actively addressing the Greek financial crisis, in early 2015,
a video was posted on Youtube\textsuperscript{TM} with the greek minister of economy, Yanis Varoufakis,
apparently ``sticking the middle finger'' at Germany to underscore his disappointment about the proposed EU economic recipes.
The video become immediately a diplomatic case.
Although minister Varoufakis quickly denounced the video as a fake, doubts persisted over its real nature,
as it was impossible to discover clear signs of manipulations.
The case became even more complicated when two more versions of the same video appeared\footnote{see http://henryjenkins.org/2015/08/f-for-fake-in-the-second-order-yanis-varoufakis-the-germans-and-the-middle-finger-that-wasnt-there.html for a full account.},
one with the minister's arm down, and another one with raised arm but two fingers sticking in a victory sign.
Frames extracted from the three videos are shown in Fig.\ref{fig:Varoufakis_Frames}.
Obviously, at least two of the videos had been manipulated.
We therefore applied our algorithm to the videos in search of clues of what really happened.
Although the videos were several minutes long, the sequence with the raised finger, where the videos differ, lasted just a few seconds,
so we could adopt an asymmetric modality of analysis,
focusing only to this section and looking for possible matching in the rest of the video.
This circumstance made the computational effort fully acceptable.

\begin{figure}	
	\begin{tabular}{cc}
		\includegraphics[scale=0.85]{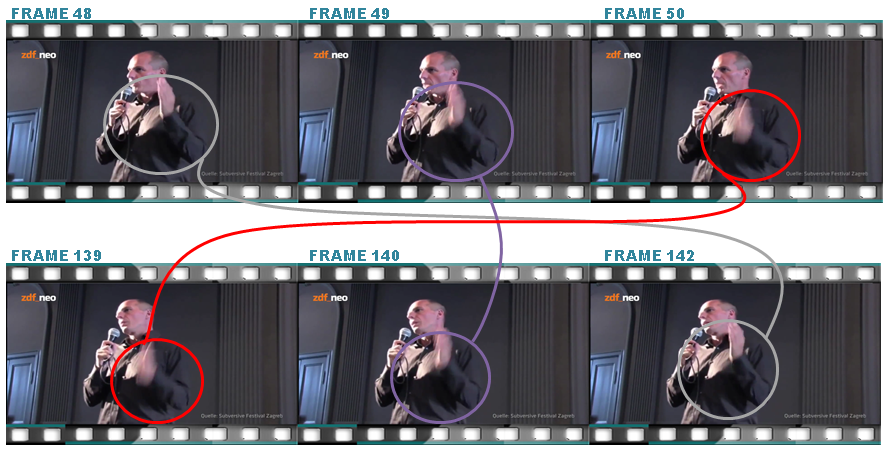} \\
		{\footnotesize (a) Matching frames} \\
		\includegraphics[scale=0.85]{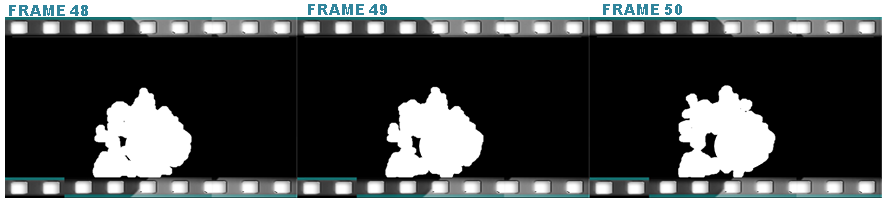} \\
		{\footnotesize (b) Detection maps}
	\end{tabular}
	\caption{Findings in the Varoufakis video \#2 (arm down). Top, evidence of copy-move with flipping. Bottom, sample detection maps.}
\label{fig:Varoufakis_Maps}
\end{figure}

The proposed algorithm did not discover any copy-moves in videos \#1 (middle-finger) and \#3 (victory).
Since only one of them (at most) can be pristine, we are missing a forgery.
A first possible explanation is that the victory video is original, and the other one is obtained by hiding the index finger through inpainting, very easy on small areas.
However, it is also possible that the middle-finger video is original, and the victory sign is obtained by copy-moving the index from somewhere else.
However, for such a small copy-move detection becomes very unlikely for any algorithm.
The proposed algorithm was instead able to detect a clear forgery in video \#2 (arm down),
a copy-move with flipping from a temporally close section of the same video.
Fig.\ref{fig:Varoufakis_Maps} shows the relevant frames, with the matching regions, and the corresponding detection maps.
{\red
To obtain a visual confirmation of this finding, we played two instances of video \#2 side by side, one going forward and the other backward in time.
With suitable synchronization, the copy-move appeared obvious, and could easily pass the scrutiny of a court of justice.}
Therefore, the proposed method seems to work also outside the laboratory,
barring prohibitive conditions where any algorithm would fail.

\section{Conclusion}

We have proposed a method for the detection and localization of video copy-moves.
Since keypoint-based approaches are ineffective with most occlusive forgeries, we focused on dense-field methods.
With this approach, the main issue is complexity, especially for videos, cursed by their huge data size.
To deal with this problem we resorted to a fast randomized patch matching algorithm, a hierarchical analysis strategy, and parallel implementation.
Experiments confirm that the proposed method has an excellent detection and localization ability,
also for occlusive copy-moves, and even in adverse scenarios including rotated copy-moves and compressed videos.
Moreover, the running time is much reduced w.r.t. linear search, enabling practical video analysis.

Despite all efforts, the proposed method cannot be used for real-time analysis or mass screening of video repositories.
Therefore, there is much room for future research on tools that solve these problems, even at the price of reduced reliability.
We ourselves are currently working on the development of fast keypoint-based methods for video analysis.

\section*{Acknowledgment}

This material is based on research sponsored by the Air Force Research Laboratory
and the Defense Advanced Research Projects Agency under agreement number FA8750-16-2-0204.
The U.S.Government is authorized to reproduce and distribute reprints for Governmental purposes
notwithstanding any copyright notation thereon.
The views and conclusions contained herein are those of the authors
and should not be interpreted as necessarily representing the official policies or endorsements,
either expressed or implied, of the Air Force Research Laboratory
and the Defense Advanced Research Projects Agency or the U.S. Government.

\balance

\bibliographystyle{IEEEbib}
\bibliography{forensic}

\begin{thebibliography}{10}

\bibitem{Milani2012}
S.~Milani, M.~Fontani, P.~Bestagini, M.~Barni, A.~Piva, M.~Tagliasacchi, and
  S.~Tubaro,
\newblock ``An overview on video forensics,''
\newblock {\em APSIPA Transactions on Signal and Information Processing}, vol.
  1, December 2012.

\bibitem{Stamm2012}
M.C. Stamm, W.S. Lin, and K.J.~Ray Liu,
\newblock ``Temporal forensics and anti-forensics for motion compensated
  video,''
\newblock {\em IEEE Transactions on Information Forensics and Security}, vol.
  7, no. 4, pp. 1315--1329, August 2012.

\bibitem{Wang2006}
W.~Wang and H.~Farid,
\newblock ``Exposing digital forgeries in video by detecting double {MPEG}
  compression,''
\newblock in {\em ACM Workshop on Multimedia and Security}, 2006, pp. 37--47.

\bibitem{Gironi2014}
A.~Gironi, M.~Fontani, T.~Bianchi, A.~Piva, and M.~Barni,
\newblock ``A video forensic technique for detection frame deletion and
  insertion,''
\newblock in {\em IEEE International Conference on Acoustics, Speech and Signal
  Processing}, May 2014, pp. 6226--6230.

\bibitem{Su2009}
Y.~Su, J.~Zhang, and J.~Liu,
\newblock ``Exposing digital video forgery by detecting motion-compensated edge
  artifact,''
\newblock in {\em International Conference on Computational Intelligence and
  Software Engineering}, 2009, pp. 1--4.

\bibitem{Wu2014}
Y.~Wu, X.~Jiang, T.~Sun, and W.~Wang,
\newblock ``Exposing video inter-frame forgery based on velocity field
  consistency,''
\newblock in {\em IEEE International Conference on Acoustics, Speech and Signal
  Processing}, 2014, pp. 2674--2678.

\bibitem{Feng2014}
C.~Feng, Z.~Xu, W.~Zhang, and Y.~Xu,
\newblock ``Automatic location of frame deletion point for digital video
  forensics,''
\newblock in {\em ACM workshop on Information hiding and multimedia security},
  2014, pp. 171--179.

\bibitem{Ravi2014}
H.~Ravi, A.V. Subramanyam, G.~Gupta, and B.~Avinash Kumar,
\newblock ``Compression noise based video forgery detection,''
\newblock in {\em IEEE International Conference on Image Processing}, 2014, pp.
  5352--5356.

\bibitem{Wickramasuriya2005}
J.~Wickramasuriya, M.~Alhazzazi, M.~Datt, S.~Mehrotra, and
  N.~Venkatasubramanian,
\newblock ``Privacy-protecting video surveillance,''
\newblock in {\em SPIE Int.l Symposium on Electronic Imaging}, 2005, pp.
  64--75.

\bibitem{Granados2012a}
M.~Granados, J.~Tompkin, K.~Kim, O.~Grau, J.~Kautz, and C.~Theobalt,
\newblock ``How not to be seen ­ object removal from videos of crowded
  scene,''
\newblock in {\em Computer Graphics Forum 31}, 2012, pp. 219--228.

\bibitem{Granados2012b}
M.~Granados, K.~Kim, J.~Tompkin, J.~Kautz, and C.~Theobalt,
\newblock ``Background inpainting for videos with dynamic objects and a
  free-moving camera,''
\newblock in {\em European Conference on Computer Vision (ECCV)}, 2012, pp.
  682--695.

\bibitem{Wang2009}
W.~Wang and H.~Farid,
\newblock ``Exposing digital forgeries in video by detecting double
  quantization,''
\newblock in {\em ACM Workshop on Multimedia and Security}, 2009, pp. 39--48.

\bibitem{Sun2012}
T.~Sun, W.~Wang, and X.~Jiang,
\newblock ``Exposing video forgeries by detecting {MPEG} double compression,''
\newblock in {\em IEEE International Conference on Acoustics, Speech, and
  Signal Processing}, 2012, pp. 1389--1392.

\bibitem{Labartino2013}
D.~Labartino, T.~Bianchi, A.~De Rosa, M.~Fontani, D.~Vazquez-Padin, A.~Piva,
  and M.~Barni,
\newblock ``Localization of forgeries in {MPEG}-2 video through {GOP} size and
  {DQ} analysis,''
\newblock in {\em IEEE International Workshop on Multimedia Signal Processing},
  2013, pp. 494--499.

\bibitem{He2016}
P.~He, X.~Jiang, T.~Sun, and S.~Wang,
\newblock ``Double compression detection based on local motion vector field
  analysis in static-background videos,''
\newblock {\em Journal of Visual Communication and Image Representation}, vol.
  35, pp. 55--66, 2016.

\bibitem{Chen2008}
M.~Chen, J.~Fridrich, M.~Goljan, and J.~Luk{\'{a}}s,
\newblock ``Determining image origin and integrity using sensor noise,''
\newblock {\em IEEE Transactions on Information Forensics and Security}, vol.
  3, no. 1, pp. 74--90, March 2008.

\bibitem{Chierchia2014}
G.~Chierchia, G.~Poggi, C.~Sansone, and L.~Verdoliva,
\newblock ``A {Bayesian}-{MRF} approach for {PRNU}-based image forgery
  detection,''
\newblock {\em IEEE Transactions on Information Forensics and Security}, vol.
  9, no. 4, pp. 554--567, April 2014.

\bibitem{Mondaini2007}
N.~Mondaini, R.~Caldelli, A.~Piva, M.~Barni, and V.~Cappellini,
\newblock ``Detection of malevolent changes in digital video for forensic
  applications,''
\newblock in {\em Proc. of SPIE Conference on Security, Steganography and
  Watermarking of Multimedia}, 2007, vol. 6505.

\bibitem{Hsu2008}
C.-C. Hsu, T.-Y. Hung, C.-W. Lin, and C.-T. Hsu,
\newblock ``Video forgery detection using correlation of noise residue,''
\newblock in {\em IEEE International Workshop on Multimedia Signal Processing},
  2008, pp. 170--174.

\bibitem{Chen2015}
S.~Chen, S.~Tan, B.~Li, and J.~Huang,
\newblock ``Automatic detection of object-based forgery in advanced video,''
\newblock {\em IEEE Transactions on Circuits and Systems for Video Technology},
  in press 2015.

\bibitem{Davino2017}
D.~D'Avino, D.~Cozzolino, G.~Poggi, and L.~Verdoliva,
\newblock ``Autoencoder with recurrent neural networks for video forgery
  detection,''
\newblock in {\em IS\&T International Symposium on Electronic Imaging: Media
  Watermarking, Security, and Forensics}, 2017.

\bibitem{Kobayashi2010}
M.~Kobayashi, T.~Okabe, and Y.~Sato,
\newblock ``Detecting forgery from static-scene video based on inconsistency in
  noise level functions,''
\newblock {\em IEEE Transactions on Information Forensics and Security}, vol.
  5, no. 4, pp. 883--892, December 2010.

\bibitem{Wang2007}
W.~Wang and H.~Farid,
\newblock ``Exposing digital forgeries in video by detecting duplication,''
\newblock in {\em ACM Multimedia and Security Workshop}, 2007, pp. 35--42.

\bibitem{Bestagini2013}
P.~Bestagini, S.~Milani, M.~Tagliasacchi, and S.~Tubaro,
\newblock ``Local tampering detection in video sequences,''
\newblock in {\em IEEE International Workshop on Multimedia Signal Processing},
  October 2013, pp. 488--493.

\bibitem{Liao2013}
S.-Y. Liao and T.-Q. Huang,
\newblock ``{Video copy-move forgery detection and localization based on Tamura
  texture features},''
\newblock in {\em International Congress on Image and Signal Processing}, 2013,
  pp. 864--868.

\bibitem{Subramanyam2012}
A.~Subramanyam and S.~Emmanuel,
\newblock ``Video forgery detection using {HOG} features and compression
  properties,''
\newblock in {\em IEEE International Workshop on Multimedia Signal Processing},
  2012, pp. 89--94.

\bibitem{Cozzolino2015}
D.~Cozzolino, G.~Poggi, and L.~Verdoliva,
\newblock ``Efficient dense-field copy-move forgery detection,''
\newblock {\em IEEE Transactions on Information Forensics and Security}, vol.
  10, no. 11, pp. 2284--2297, November 2015.

\bibitem{Damiano2015}
L.~D'Amiano, D.~Cozzolino, G.Poggi, and L.~Verdoliva,
\newblock ``{Video forgery detection and localization based on {3D}
  PatchMatch},''
\newblock in {\em IEEE International Conference on Multimedia and Expo
  Workshops}, 2015, pp. 1--6.

\bibitem{Barnes2009}
C.~Barnes, E.~Shechtman, A.~Finkelstein, and D.B. Goldman,
\newblock ``Patchmatch: a randomized correspondence algorithm for structural
  image editing,''
\newblock {\em ACM Transactions on Graphics}, vol. 28, no. 3, 2009.

\bibitem{Barnes2010}
C.~Barnes, E.~Shechtman, D.B. Goldman, and A.~Finkelstein,
\newblock ``The generalized patchmatch correspondence algorithm,''
\newblock in {\em European Conf. on Computer Vision}, 2010, vol. 6313, pp.
  29--43.

\bibitem{Cozzolino2014}
D.~Cozzolino, G.~Poggi, and L.~Verdoliva,
\newblock ``Copy-move forgery detection based on {PatchMatch},''
\newblock in {\em IEEE International Conf. on Image Processing}, 2014, pp.
  5312--5316.

\bibitem{Christlein2012}
V.~Christlein, C.~Riess, J.~Jordan, and E.~Angelopoulou,
\newblock ``An evaluation of popular copy-move forgery detection approaches,''
\newblock {\em IEEE Transactions on Information Forensics and Security}, vol.
  7, no. 6, pp. 1841--1854, 2012.

\bibitem{Pan2010}
X.~Pan and S.~Lyu,
\newblock ``Region duplication detection using image feature matching,''
\newblock {\em IEEE Transactions on Information Forensics and Security}, vol.
  5, no. 4, pp. 857--867, December 2010.

\bibitem{Amerini2011}
I.~Amerini, L.~Ballan, R.~Caldelli, A.~Del Bimbo, and G.~Serra,
\newblock ``A {SIFT}-based forensic method for copy–move attack detection and
  transformation recovery,''
\newblock {\em IEEE Transactions on Information Forensics and Security}, vol.
  6, no. 3, pp. 1099--1110, 2011.

\bibitem{Zhao2013}
J.~Zhao and W.~Zha,
\newblock ``Passive forensics for region duplication image forgery based on
  {Harris} feature points and local binary patterns,''
\newblock in {\em Mathematical Problems in Engineering}, 2013, pp. 1--12.

\bibitem{Fridrich2003}
J.~Fridrich, D.~Soukal, and J.~Luk{\'{a}}s,
\newblock ``Detection of copy-move forgery in digital images,''
\newblock in {\em proc. of Digital Forensic Research Workshop}, 2003.

\bibitem{Muhammada2012}
G.~Muhammada, M.~Hussain, and G.~Bebis,
\newblock ``Passive copy move image forgery detection using undecimated dyadic
  wavelet transform,''
\newblock {\em Digital Investigation}, vol. 9, pp. 49--57, 2012.

\bibitem{Mahdian2007}
B.~Mahdian and S.~Saic,
\newblock ``Detection of copy–move forgery using a method based on blur
  moment invariants,''
\newblock {\em Forensic Science International}, vol. 171, pp. 180--189, 2007.

\bibitem{Zhao2013a}
J.~Zhao and J.~Guo,
\newblock ``Passive forensics for copy-move image forgery using a method based
  on {DCT} and {SVD},''
\newblock {\em Forensic Science International}, vol. 233, pp. 158--166, 2013.

\bibitem{Ryu2013}
S.-J. Ryu, M.~Kirchner, M.-J. Lee, and H.-K. Lee,
\newblock ``Rotation invariant localization of duplicated image regions based
  on {Zernike} moments,''
\newblock {\em IEEE Transactions on Information Forensics and Security}, vol.
  8, no. 8, pp. 1355--1370, August 2013.

\bibitem{Li2013}
Y.~Li,
\newblock ``Image copy-move forgery detection based on polar cosine transform
  and approximate nearest neighbor searching,''
\newblock {\em Forensic Science International}, vol. 224, pp. 59--67, 2013.

\bibitem{Li2014}
L.~Li, S.~Li, H.~Zhu, and X.~Wub,
\newblock ``Detecting copy-move forgery under affine transforms for image
  forensics,''
\newblock {\em Computers and Electrical Engineering}, vol. in press, 2014.

\bibitem{Bayram2009}
S.~Bayram, H.~Sencar, and N.~Memon,
\newblock ``An efficient and robust method for detecting copy-move forgery,''
\newblock in {\em IEEE International Conference on Acoustics, Speech, and
  Signal Processing}, April 2009, pp. 1053--1056.

\bibitem{Wu2011}
Q.~Wu, S.~Wang, and X.~Zhang,
\newblock ``Log-polar based scheme for revealing duplicated regions in digital
  images,''
\newblock {\em IEEE Signal Processing Letters}, vol. 18, no. 10, pp. 559--652,
  2011.

\bibitem{Langille2006}
A.~Langille and M.~Gong,
\newblock ``An efficient match-based duplication detection algorithm,''
\newblock in {\em Canadian Conf. on Computer and Robot Vision}, 2006.

\bibitem{Teague1980}
M.R. Teague,
\newblock ``Image analysis via the general theory of moments,''
\newblock {\em Opt. Soc. Amer.}, vol. 70, no. 8, pp. 920--930, 1980.

\bibitem{Bleyer2011}
M.~Bleyer, C.~Rhemann, and C.~Rother,
\newblock ``{PatchMatch} stereo - stereo matching with slanted support
  windows,''
\newblock in {\em British Machine Vision Conference}, 2011, pp. 1--11.

\bibitem{Newson2013}
A.~Newson, A.~Almansa, M.~Fradet, Y.~Gousseau, and P.~Perez,
\newblock ``Towards fast, generic video inpainting,''
\newblock in {\em Proceedings of the 10th European Conference on Visual Media
  Production}, 2013.

\end{thebibliography}

\end{document}